\documentclass[10.5pt,letterpaper]{article}

\usepackage{authblk}

\usepackage{fullpage}

\usepackage[utf8]{inputenc} \usepackage[T1]{fontenc}    \usepackage{hyperref}       \usepackage{url}            \usepackage{booktabs}       \usepackage{amsfonts}       \usepackage{nicefrac}       \usepackage{microtype}      \usepackage{algorithm}
\usepackage{algorithmic}
\usepackage{amsmath}
\usepackage{times}
\usepackage{mathtools}

\usepackage{color}
\usepackage{multirow}
\usepackage{amsthm}
\usepackage{mathrsfs}
\usepackage{graphicx}
\usepackage{caption}\usepackage{xspace}
\usepackage{float}
\usepackage{wrapfig}
\usepackage{enumitem}
\usepackage{listings}
\usepackage{multirow}
\usepackage{bbm}
\usepackage[export]{adjustbox}
\usepackage{subcaption}

\mathchardef\mhyphen="2D

\newcommand{\vertiii}[1]{{\left\vert\kern-0.25ex\left\vert\kern-0.25ex\left\vert #1
    \right\vert\kern-0.25ex\right\vert\kern-0.25ex\right\vert}}

\def\0{{\mathbf{0}}}

\def\bbE{{\mathbb{E}}}

\def\cD{\mathcal{D}}

\def\cX{\mathcal{X}}

\def\sfX{\mathsf{X}}

\newtheorem*{rep@theorem}{\rep@title}
\newcommand{\newreptheorem}[2]{%
\newenvironment{rep#1}[1]{%
 \def\rep@title{#2 \ref{##1}}%
 \begin{rep@theorem}}%
 {\end{rep@theorem}}}
\newreptheorem{lemma}{Lemma'}
\newreptheorem{definition}{Definition'}
\newreptheorem{proposition}{Proposition'}
\newreptheorem{theorem}{Theorem'}

\newtheorem{theorem}{Theorem}
\newtheorem{lemma}[theorem]{Lemma}
\renewcommand{\text}[1]{{\textnormal{#1}}}

\usepackage{xcolor}

\usepackage{footnote}
\makesavenoteenv{tabular}
\newcommand{\Pd}{P_{X}}
\newcommand{\Pdj}{P_{X_j}}
\newcommand{\Pm}{P_{model}}

\DeclareMathOperator*{\argmax}{arg\,max}
\usepackage{natbib}

\begin{document}

\title{Competitive Training of Mixtures of Independent Deep Generative Models}

\author[1,2]{Francesco Locatello}
\author[3]{Damien Vincent}
\author[3]{Ilya Tolstikhin}
\author[1]{Gunnar R\"atsch}
\author[3]{Sylvain Gelly}
\author[2]{Bernhard Sch\"olkopf}

\affil[1]{Max-Planck Institute for Intelligent Systems, Germany}
\affil[2]{BMI, Dept. for Computer Science, ETH Zurich, Switzerland.}
\affil[3]{Google AI, Brain Team}
\date{}

{\setlength{\parindent}{0cm}
{\setlength{\parskip}{2mm}

\maketitle

\begin{abstract}
A common assumption in causal modeling posits that the data is generated by a set of independent mechanisms, and algorithms should aim to recover this structure. Standard unsupervised learning, however, is often concerned with training a single model to capture the overall distribution or aspects thereof. Inspired by clustering approaches, we consider mixtures of implicit generative models that ``disentangle'' the independent generative mechanisms underlying the data. Relying on an additional set of discriminators, we propose a competitive training procedure in which the models only need to capture the portion of the data distribution from which they can produce realistic samples. As a by-product, each model is simpler and faster to train. We empirically show that our approach splits the training distribution in a sensible way and increases the quality of the generated samples.
\end{abstract}
\section{Introduction}
In recent years, (implicit) generative models have attracted significant attention in machine learning. Generative models are trained from unlabelled data and are capable of generating samples which resemble the ones from the training distribution. This task is considered of crucial importance for unsupervised learning.
Two of the most prominent approaches are Generative Adversarial Networks (GANs)~\citep{goodfellow2014generative} and Variational Autoencoders (VAEs)~\citep{kingma2013auto}. Both approaches aim at minimizing the discrepancy between the true data distribution and the one learned by the model. The model distribution is typically parametrized with a neural network which transforms random vectors into points in the space of the training data (e.g., images). Variational Autencoders maximize a log-likelihood and are able to perform efficient approximate inference on probabilistic models with continuous latent variables and an intractable posterior. Furthermore, they come with an encoder network which maps data points to the latent space. Unfortunately, VAEs are known to produce blurry samples when applied to natural images. GANs take a completely different approach, relying on adversarial training. This yields impressive empirical results; however, adversarial training comes at a cost. GANs are harder to train and suffer from the \textit{mode collapse} problem. If the data distribution lies outside the class of functions that the generator network can effectively learn, the network will ignore portions of the data and focus on the parts which can be approximated well given its limited capacity. A number of approaches have been presented to tackle this problem. The most relevant to our setting is the work of~\cite{Tolstikhin:2017wo} who proposed to train a sequence of multiple generators which are subsequently mixed. As long as each generator collapses on a different mode, and given a large enough number of generators, one can thus approximate the whole data distribution by combining them. Several works followed up, trying to avoid the sequential training of the generators. In particular, \cite{hoang2017multi} proposed to train multiple generators in parallel using adversarial training, and a classifier to help them specialize to different modes.
In contrast to GANs, VAEs are not trained with a minimax game and always model the whole support of the data distribution. Arguably, both extremes are undesirable. While approximating the whole distribution is the aim of generative density estimation, it should not come at the cost that the approximation is too poor to be useful.

Inspired by the causal interpretation of generative models, we aim at bridging this gap by developing a general approach to train multiple models in parallel which focus on independent parts of the training distribution. As a consequence, each generative model will be able to collapse on some modes while the mixture of generators will still approximate the whole data distribution. Our approach lies in between VAEs and GANs and is rooted in the causal modeling problem of disentangling independent generative mechanisms. We introduce an additional set of discriminators which act as observers of the training, influencing the sampling distribution of the optimizer so that different generative models can specialize on different parts of the data distribution.  

In this paper we design a competitive training procedure to split the data distribution into components which are approximated well by independent generative models. Our contributions can be summarized as:
\begin{itemize}
\item We provide an algorithmic framework for training mixtures of generative models. We target the most abstract case of minimizing a general $f$-divergence using a mixture of generators which can be trained in parallel. 
\item We instantiate our framework to mixtures of VAEs and GANs. By doing so, we allow them to collapse on separate modes of the distribution. As a consequence, the models no longer need to cover all modes of the possibly complex overall distribution, and can thus use their limited capacity to produce realistic samples within the support of their portion of the data distribution. 
\item We relate our training procedure with clustering approaches, showing that our model recovers k-means as a special case.
\item We provide empirical evidence that shows that our training procedure splits the training distribution into distinct components, significantly increasing the log-likelihood estimate for synthetic data and improving the FID score on MNIST and celebA.
\end{itemize}
\vspace{-0mm}
\section{(Causal) Generative Mixtures: Problem Setting}
Let $\cD_X$ be a dataset composed of $N$ samples $x_i$ from $X$. 
Furthermore, let $\Pd$ be an unknown data distribution defined over the data space $\cX$ with support $\sfX$ to be approximated with an easy to sample distribution $\Pm$. Given a probability distribution $P$, we denote $dP$ its density. The goal of implicit generative density estimation is to make the samples from both distributions $\Pd$ and $\Pm$ look alike. This is typically formulated as some optimization problem minimizing the disagreement between the two.
To measure such disagreement it is common practice to use an $f$-divergence:
\begin{align}\label{eq:f-Div}
D_f(Q\|P):= \int_{x\in\cX} f\left(\frac{dQ}{dP}(x)\right) dP(x)
\end{align}
where $f(1)=0$ and $f$ is convex. 
The goal is then to solve the following optimization problem:
\begin{align*}
\min_{\Pm} D_f(\Pm\|\Pd)
\end{align*}
Unfortunately, $\Pd$ is unknown and only an empirical estimate of $D_f$ is available through the samples in the training set $\cD_X$.
While this setting is at the heart of the adversarial training of GANs \citep{goodfellow2014generative}, VAEs~\citep{kingma2013auto} minimize a variational bound on $D^{KL}(\Pd\|\Pm)$ which is one of the various divergences which can be written in the form of~\eqref{eq:f-Div} with the appropriate choice of $f$~\citep{nowozin2016f}.

Mixture of generative models are well motivated in the literature on causality. We assume that the data was generated by \textit{independent mechanisms}, i.e., that the generative process of the overall distribution is composed of separate modules that do not inform nor influence each other~\citep{PetJanSch17}. We aim at modeling each of these mechanisms with an implicit generative model.
Consider the special case of a variable $X$ which is caused by (mixing) several independent sources $g_1,\dots,g_K$ without parents in the causal graph. Then, generative models take the form \citep{SpiGlySch93,Pearl00} 
\begin{equation}\label{eq:markov2}
p({X,g_1,\dots,g_K}) = p(X | g_1,\dots,g_K) \prod_{j = 1}^K p(g_j).
\end{equation}
The terms on the right hand side are referred to as {\em causal conditionals, Markov kernels}, or {\em mechanisms}. 
Note that only one of the mechanisms, $p(X | g_1,\dots,g_K)$, implementing the mixing, is a conditional; the others reduce to unconditional distributions since the sources have no parents in the causal graph. The conditional can be written as a structural equation \citep{Pearl00} 
\begin{equation}\label{eq:SEM}
X := f(g_1,\dots,g_K,c) \equiv g_c,
\end{equation}
where $c$ is a discrete noise variable taking values in $\{1,\dots,K\}$. The distribution of $c$ determines the mixing coefficients. Eq.\ (\ref{eq:SEM}) expresses the conditional as a mechanism represented by a noisy function.

It has been argued that the true causal factorization is the simplest among all the possible factorizations of the random variables, in the sense that the sum of the complexities of the causal conditionals is minimized~\citep{janzing2010causal}. 
Suppose each training point was generated by one of the mechanisms $g_j$, but we get to observe only the mixture of all these realizations. Recovering the mechanisms is then a hard and ill-posed inverse problem, since there are many ways to represent the same mixture in terms of the different components. Solving this problem amounts to learning a particular kind of structural causal generative model, and it could form a building block of more complex causal models \citep{SchJanLop16}.

We make the simplifying assumption that the supports of the different generative mechanisms do not overlap, hence if we observe two identical realizations of $X$, they must have been generated by the same mechanism. While a soft assignment is also possible, we focus on training with hard assignments which is known to converge faster~\citep{kearns1998information}. From a practical perspective, this implies dividing the data distribution into non-overlapping components, each approximated by an independent generative model. 
This way, we allow each generative model to learn only a fraction of the data distribution. As a consequence, they are able to model each part of the distribution better than a single model trying to approximate the whole distribution. While the optimal split of the data distribution is unknown, we rely on a competitive procedure based on a set of discriminators. Each training point will be {\em won} by the model which generates the most similar samples according to the set of discriminators. These discriminators can only influence the distribution of the training signal of the generative models and are not used to propagate explicit gradients.
Intuitively, a model that approximates the true causal mechanism will be easier to learn, and hence will generate better samples. 

\looseness=-1We can think of the generative process as an ``entanglement'' of independent mechanisms~\citep{janzing2010causal}.\footnote{Here, independence refers to physical processes that do not influence or inform each other, i.e., knowing or changing one process does not affect (knowledge of) another process.}
Therefore, to better approximate the data distribution one can consider $\Pm$ as a mixture of distributions:
\begin{align}\label{eq:mixture}
\Pm=\sum_{j=1}^k\alpha_jP_{g_j}
\end{align}
each of them specialized on one of the generating mechanisms. Training mixtures of experts with boosting algorithms like the one in \citep{Tolstikhin:2017wo} has favorable optimization properties. 
In particular, adding components to a mixture is a convex optimization problem~\citep{LocDreKhaValRat18,LocKhaGhoRat18}. On the other hand, in the context of deep learning, sequential training comes at a great cost in time. The sequential nature of boosting-like algorithms requires that each model is fully trained before the subsequent ones begin training, and an already trained model is never changed afterwards. Subsequent models are trained to fit the parts of space that the previous mixture could not approximate well. As a consequence, there is no incentive for any of the models to focus on a mode. Each generator will try to cover the whole residual of the data distribution which is not yet approximated. 
In contrast to \citep{Tolstikhin:2017wo}, rather than training the mixture of generative models sequentially, we train them at the same time. By doing so, we lose the convexity of the objective; however, if one is able to decouple the training procedure, each model can be trained in parallel. Furthermore, instead of relying on mode collapse to happen, we let the models compete for training points in order to force them to learn different parts of the data distribution.

\looseness=-1Finally, disentangling independent causal generative models avoids the problems expressed by Theorem 1 in~\citep{locatello2018challenging}. While the notion of disentanglement discussed in that paper differs from ours (disentangling factors of variation as opposed to the generative mechanisms), their analysis would apply to our setting without the assumption of independent causal mechanisms of Equation~\eqref{eq:SEM} which explicitly introduce an inductive bias we leverage for training. 

\vspace{-0mm}
\section{Training Independent Generative Models}
Borrowing ideas from the clustering literature, at each iteration, each generative model $g_j$ is trained on a different portion of the dataset.
We assign each training point $x_i$ to a single model using a set of $K$ binary partition functions $c_j$ implementing the realization of the mixing distribution $c(x_i)$ (i.e. $c_j(x_i) = 1$ and $c_{[K]\setminus j}(x_i) = 0$ when $c(x_i) = j$). 
Intuitively, our training procedure is related to the k-means algorithm. In k-means, one first decouples the training data across the centroids and then updates the centroids based on the assignment. 
Our approach is outlined in Algorithm~\ref{algo:informal_algo}.
\begin{algorithm}[H]
\begin{algorithmic}[1]
  \STATE \textbf{init} $K$ generative models $g_j$, $c^{(0)}_j$, $j = 1,\ldots, K$
  \STATE \textbf{for} {$t=0\dots T$}
  \STATE \quad  $\min_{P_{g_j}} D_f\left(P_{g_j}\|\Pdj^{(t)}\right)$ for every $g_j$
  \STATE \quad Update $c_j^{(t+1)}(x)$ for every $g_j$
  \STATE \textbf{end for}
\end{algorithmic}
 \caption{Mixture training}\label{algo:informal_algo}
\end{algorithm}
Formally, let us consider $K$ assignment functions $c_j(x): \cX \rightarrow \left\lbrace 0,1\right\rbrace$.
We assume that for any $x\in\cX$ there is a unique $j$ such that $c_j(x) = 1$ and it is zero for all the others $c_{[K]\setminus j}$.
Let us now use the $c_j$ to partition the support $\sfX$ of $d\Pd$.
First of all, let us define:
\begin{align*}
d\Pdj (x) := \begin{cases} \frac{d\Pd (x)}{\int_\sfX d\Pd (x)c_j(x)}\text{,} & \mbox{if } c_j(x) = 1 \\ 0, & \mbox{otherwise. }\end{cases}
\end{align*}
In Algorithm~\ref{algo:informal_algo}, when the assignment function is some fixed $c_j^{(t)}$, we write $d\Pdj^{(t)}$ to make the dependency on the particular assignment explicit.  
For a given $c_j$, we define the mixing proportions $\alpha_j$ of $\Pm$ as the normalization constant of $\Pdj$. This can be empirically estimated by counting how many training points are assigned to the $j$-th generator.
We now show how to decouple the training of the generators by minimizing an upper bound of the $f$-divergence.

\begin{lemma}\label{lemma:parallel_training}
For a fixed partition function $c_j^{(t)}$, we minimize for all $j \in [K]$:
\begin{align}\label{eq:objective}
\min_{P_{g_j} } \sum_j \alpha_j D_f(P_{g_j}\| \Pdj^{(t)}) ,
\end{align}
which is an upper bound on the $f$-divergence for a mixture model. We defer the proof to Appendix~\ref{lemma_app:parallel_training}
\end{lemma}

Since each term in the sum in Equation~\eqref{eq:objective} is independent, each generative model can be trained independently to approximate $\Pdj^{(t)}$.

After training the generative models, we fix them and update the assignment of each training point.
Intuitively, our goal is that the mixture of $P_{g_j}$ resembles as much as possible $\Pd$. Therefore, at iteration $t$, we first compute the likelihood $P_{g_j}^{(t)}(x_i)$ of each training point $x_i$. Then, we update the partition function $c_{j}^{(t+1)}(x_i)$ assigning each training point to the maximum likelihood model. 
  
We estimate the likelihood by training a discriminator to distinguish samples from $P_{g_j}$ and samples from $\Pd$. Let $D_{g_j}(x)$ be the output of the $j$-th discriminator. Then:
\begin{align}\label{eq:classifier}
D^{(t)}_{g_j}(x) \approx \frac{ d\Pd(x)}{  d\Pd(x) + dP_{g_j}^{(t)}(x)}.
\end{align}
After training the classifier, we can rewrite Equation~\eqref{eq:classifier} as:
\begin{align*}
dP_{g_j}^{(t)}(x) \approx d\Pd(x)  \frac{1-D^{(t)}_{g_j}(x)}{D^{(t)}_{g_j}(x)}
\end{align*} 
 Then, we can approximate $P_{g_j}^{(t)}(x_i)$ as the empirical estimate over the training set $\cD_X$:
\begin{align*}
P_{g_j}^{(t)}(x_i) \approx \frac{1}{Z_j}  \frac{1-D^{(t)}_{g_j}(x_i)}{D^{(t)}_{g_j}(x_i)}
\end{align*}
where $Z_j = \sum_{x\in \cD_X} \frac{1-D^{(t)}_{g_j}(x)}{D^{(t)}_{g_j}(x)}$.
We now assign each training point to the mechanism $j$ that generates the most similar samples, i.e., the probability of sampling $x_i$ from $g_j$ is larger than the one of sampling it from other mechanisms $g_{[K]\setminus j}$:
\[
   c_j^{(t+1)}(x_i) = \left\{\begin{array}{ll}
        1, & \text{if } j = \argmax_l P_{g_l}^{(t)}(x_i) \\
        0, & \text{otherwise. }
        \end{array}\right. 
  \]
  
Intuitively, if a generator $g_j$ is producing samples similar to a training point $x_i$, we will have that $P_{g_j}^{(t)}(x_i)$ large. On the other hand, if $x_i$ is different than the samples generated from $g_j$ then $P_{g_j}^{(t)}(x_i)$ is small.

A sketch of the training procedure, using VAEs decoders as generators, is depicted in Figure~\ref{fig:model}. To sample from a VAE decoder $g_j$ we sample from $P_{g_j}(x) = \int P_{g_j}(x|z) dP_z(z)$, which is an uncountable mixture of Gaussians obtained by marginalizing the decoder over the prior. Practically speaking, we decode a noise vector sampled from the prior. 

\textbf{Independence as inductive bias:}
\looseness=-1After an initial pre-training on a random uniform split of the dataset, we assume that all the models have learned to generate the low level features of images up to some degree. 
To learn independent causal generative models, we leverage the assumption that the independent mechanisms do not inform nor influence each other~\citep{PetJanSch17}. Let $x_i$ be a training point which was generated by one of the true causal mechanisms $j$. Whenever a model $g_j$ train on $x_i$, it will make some progress on learning the true generative mechanisms, therefore $D_f(P_{g_j}\| \Pdj)$ will decrease. On the other hand, due to the independence assumption $D_f(P_{g_j}\| P_{X_{[K]\setminus j}})$ will not decrease. On images, the independence assumption is somewhat restrictive. Indeed, the low level features are common across the different generative mechanisms. Our assumption is that after pre-training, the independence assumption will approximately hold. While $D_f(P_{g_j}\| P_{X_{[K]\setminus j}})$ will not exactly be constant, the improvement will be minimal compared to the one on $D_f(P_{g_j}\| \Pdj)$.
If one wants to train the K generative models from scratch without the pre-training, some form of parameter sharing (at least on the layers close to the pixel space) is fundamental. Indeed, a method training on $x_i$ will not only improve on the mechanisms $\Pdj$ but also on all the others. Therefore, a model that was intialized better than the others could in principle win all the training points. To further accommodate to the violation of the independence assumption, we enforce that each model always keeps training on its best points if it is not winning enough training data. While we keep training the model, we update its prior accordingly, so that when sampling from the mixture, the model will have no weight. By continuing its training, models which loose all points can recover and start winning training data again. We found that uniform pre-training and this form of load balancing yielded faster convergence. 

\textbf{Relation to EM:}
In the generative interpretation of clustering, one assumes that the data was generated from each centroid $\mu$ with an additive Gaussian noise vector, i.e., 
\begin{equation}\label{eq:k-means}
x = \mu + \epsilon.
\end{equation}
This formulation naturally yields an euclidean cost for the cluster assignment when decoupling the data between the different centroids. 
Unfortunately, the generative model of k-means is not powerful enough to generate images. On the other hand, we adopt the same training procedure of k-means by training more complex generative models. 
\begin{lemma}\label{lemma:k-means}
The training procedure in Algorithm~\ref{algo:informal_algo} degenerates to k-means using a degenerate generative model of the form of Equation~\ref{eq:k-means}. Such a generative model, can be obtained with a degenerate VAE.
\end{lemma}
This lemma represents the (trivial) sanity check that our algorithm is strictly more powerful than k-means and that the additional machinery we introduced with the density ratio trick does not change the essence of the training procedure. This does not come as a surprise after we show that k-means can be obtained from our framework using a degenerate VAE architecture as generative model (constant encoder with gaussian decoder where the network parameterizing the mean is the identity). Note that having a single $K+1$ classifier instead of $K$ discriminators could also be possible in practice but would break Lemma~\ref{lemma:k-means} as its output would not be an estimate of the likelihood $P_{g_j}(x_i)$.

\textbf{Consequences for generative models}
Our general framework can be instantiated for arbitrary generative models. We now discuss the advantage that our framework brings to VAEs and GANs.
First of all, VAEs do not exhibit mode-collapse. In our framework, while each model will diligently try to cover the whole support of the assigned portion of the data distribution, the different models will actively compete with each other trying to generate better and better samples. Note that while this approach is somehow inspired by GAN training, we do not directly receive gradients from the discriminators. The discriminators can only implicitly influence the generators by acting on their training distribution. In practice, it can happen that a generative model is not expressive enough to model a complex distribution. In such cases, GANs (without further regularization) will focus on just a part of the data distribution. We note that mode collapse is not only related to the capacity of the GAN, but we argue that a model prone to mode collapse due to training instability will exhibit mode collapse also when its capacity is too little. Instead, VAEs try to cover the whole data distribution as a consequence of maximum likelihood. Therefore, VAEs are overly ``inclusive'' when used as generative models and tend to produce samples outside the support of the data distribution. These \textit{bridges} between different modes can be seen in Figure~\ref{pic:blurry}. While we do not claim that the \textit{only} reason for blurriness are samples out of the support of the data distribution, we argue that it is unlikely that these samples are realistic. Regularized variants of GANs also exhibits the same issue. 
Furthermore, if one wants to generate approximate samples from the data distribution, there should be an incentive for the support of the model to be as close as possible to $\sfX$. In particular, if $\sfX$ is disjoint, $\Pm$ should also have disjoint support.

\begin{figure*}
\begin{minipage}{0.6\textwidth}
\hspace{-7mm}
\includegraphics[scale=0.6]{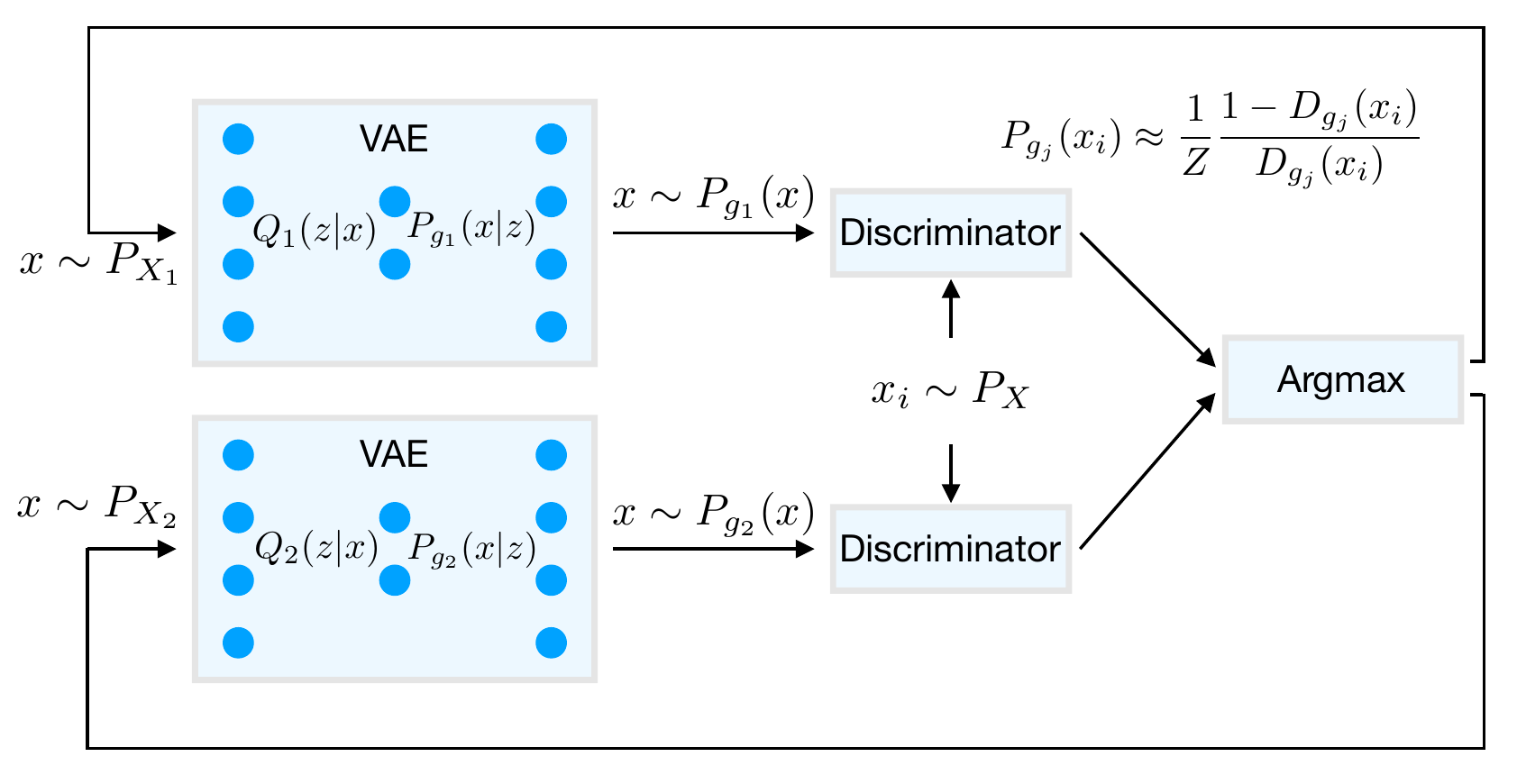}
\captionof{figure}{Training pipeline with two VAEs}\label{fig:model}
\end{minipage}
\hspace{-3mm}
\begin{minipage}{0.37\textwidth}
\includegraphics[scale=0.3]{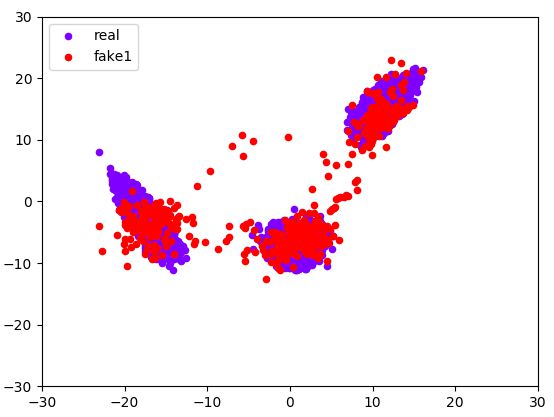}
\captionof{figure}{VAE producing samples outside the data distribution}\label{pic:blurry}
\end{minipage}
\vspace{-0mm}
\end{figure*}

\vspace{-0mm}
\section{Related Work}
\looseness=-1The main influences for our work are the literature on clustering, mixture of experts, causality, and implicit generative models.

Clustering is a cornerstone of unsupervised learning and the literature is vast, see~\citep{aggarwal2013data} for a recent overview. \cite{kummerfeld2016causal} focused on the inference in structural equation models where the target variable is hidden and accessible only through indicator variables. In order to make the problem tractable they look for indicators that only depend on the latent variable and are independent from any other variable in the causal graph transforming the structure of the learning problem to a clustering problem. Clustering for causal discovery was already explored in~\citep{matuszewski2002double}. In the opposite direction, \cite{revolle2017clustering} looked at clustering through the lens of causal inference and proposed a semi-distance based on algorithmic complexity estimates. 

Our framework is related to competition of experts, which has been used to invert independent causal mechanisms, see~\citep{parascandolo2017learning}. Mixtures of experts with a gating network trained with EM are a classical idea which was introduced by~\cite{jordan1994hierarchical}. A more recent example applied to lifelong learning is the one in~\citep{aljundi2016expert}.
\cite{kocaoglu2017causalgan} proposed to use a GAN to learn a generative model with true observational and interventional distributions for a given causal graph while~\cite{liu2016coupled} proposed to couple several GANs to learn a joint distribution of images from different domains. 

Allowing for discrete mixtures in implicit generative model is a new trend in the community which have been explored in the setting of both VAEs~\citep{dilokthanakul2016deep,jiang2016variational} and GANs~\citep{gurumurthy2017deligan}. Note that these authors consider mixtures in the latent space rather than mixtures of independent generative models. A clustering in the latent space of VAEs was recently proposed by~\cite{van2017neural}. The work of~\citet{Tolstikhin:2017wo} first introduced the idea of training multiple implicit generative models, in order to address the mode collapse problem of GANs. \cite{ghosh2017multi} proposed to train multiple generators in parallel with a single discriminator. \citet{hoang2017multi} proposed to adversarially train a system with multiple generators and a classifier which encourages the models to split using adversarial learning. In contrast to our approach, they use the classifier entropy computed on synthetic examples as a regularizer for the objective function. This requires a delicate tuning parameter which controls the ability of the networks to split. In our setting the splitting is implicit in the training procedure. Furthermore, as the classifier is trained only on synthetic data, it is not a reliable assignment function for clustering, as real data points could be in a part of the space that the generators do not cover. Training mixtures of GANs with EM was introduced by~\cite{banijamali2017generative}. They target soft assignments using a kernel to measure the similarity between the true data points and the generated ones. In contrast to our work, they do not learn the mixing components (they are assumed uniform), and they use a kernel in the pixel space rather than a discriminator. \cite{arora2017generalization} presents a technique to achieve pure equilibrium with GANs using multiple generators and discriminators which are selected through a multi-way selector implemented with a neural network. Their approach is specific to GANs and they use a single set of discriminators which provides explicit training signal for the generators. This is very different from both the boosting and the clustering rationale as there is not a notion of assignment for the training points. Therefore, they can not address the separation of the generative mechanisms in terms of training samples and can not obtain the corresponding clustering notion.

\looseness=-1
In recent years, significant effort has gone into methods for disentangling factors of variations in data~\citep{bengio2013representation}. 
The most prominent unsupervised approaches can be identified in \citep{higgins2016beta,kim2018disentangling,chen2018isolating,kumar2017variational}.
Recently, \cite{locatello2018challenging} proved that disentangling the factors of variation is impossible without inductive biases. This is orthogonal to our setting where we disentangle the generative mechanisms and we focus on sample quality as evaluation.
On the other hand, the theorem of~\citep{locatello2018challenging} would apply to our setting as well if we did not assume independence of the causal mechanisms. Causal independence (Equation~\eqref{eq:SEM}) explicitly introduce an inductive bias to the problem that breaks the symmetry of Theorem 1 in~\citep{locatello2018challenging}.
\section{Experiments}
In this section we evaluate the proposed framework. First, we test if the model can successfully learn a set of known generative mechanisms for toy data. Then, we validate that our split of the training set indeed helps the generative models producing better samples and is indeed a scalable approach to effectively increase the capacity of the model. We want to remark that to test whether our method could be used as a tool to improve the performances of generic VAEs/GANs, we used standard architectures with standard hyperparameters in all the experiments. Most of our experiments are with VAEs, but we also perform an ablation study on the capacity of the generator using GANs. All models are initialized by training for a few epochs on the whole training set. Details of the architectures and hyperparameters along with additional visual comparisons can be found in the appendix. We will release the code after publication.

\vspace{-0mm}
\subsection{Synthetic Data}
\begin{figure}
\vspace{-2mm}
\captionof{table}{Log-likelihood of the true data under the generated distribution for the synthetic data. }\label{tab:logl}
\vspace{2mm}
\center\begin{tabular}{ l c c c  }
\toprule
   & kVAEs & bag  & VAE-150\\
  \midrule
  3 modes & \textbf{-4.59} & -6.49 & -5.42\\			
  5 modes & \textbf{-2.74} & -7.7 & -5.71\\
  9 modes & \textbf{-2.51} & -7.05 & -6.83\\  
  \midrule
\end{tabular}
\vspace{-0mm}
\end{figure}

\begin{figure*}[h]
\vspace{-0mm}
\begin{minipage}{\textwidth}
\center\includegraphics[scale=0.21]{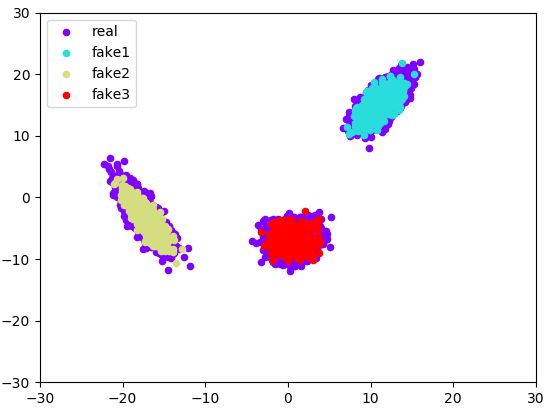}
\includegraphics[scale=0.21]{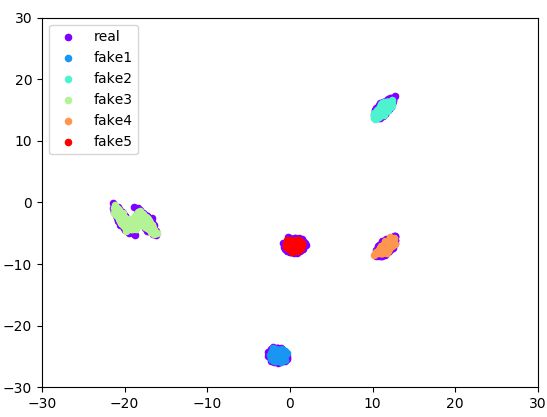}
\includegraphics[scale=0.21]{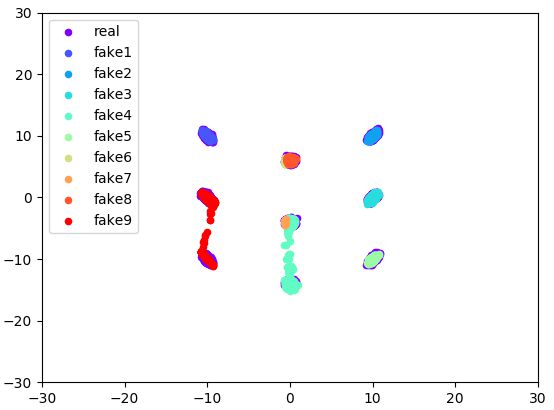}
\captionof{figure}{Synthetic data experiment with different number of modes. Our models split the data distribution and only learn simple components.}\label{pic:gmm3}
\vspace{5mm}
\begin{minipage}{\textwidth}
\center\includegraphics[scale=0.12]{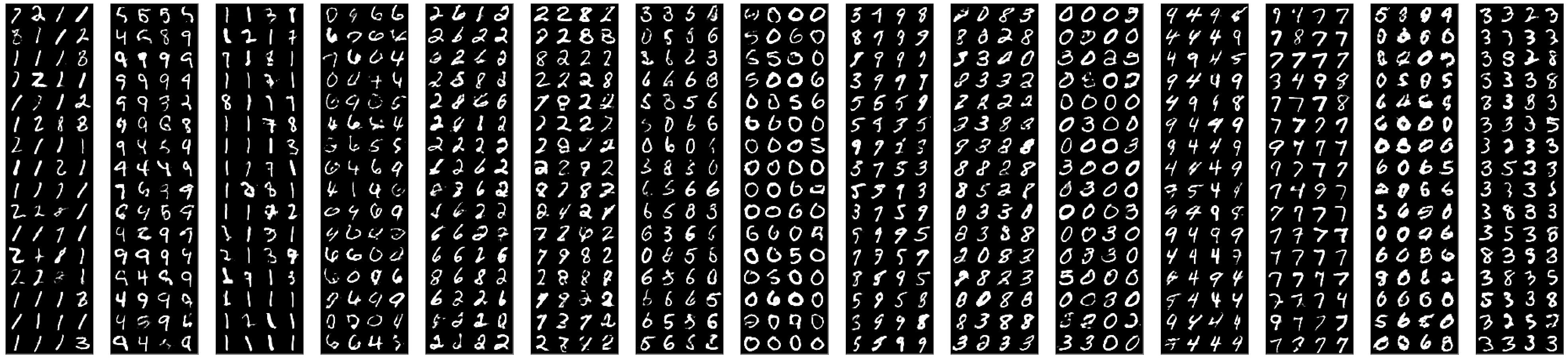}
\center\includegraphics[scale=0.12]{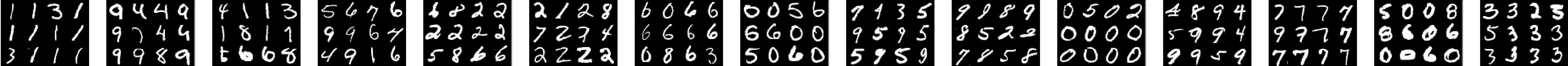}
\captionof{figure}{MNIST: samples (above) from VAEs separated per model (mixture component) and real digits clustered after just 10 iterations of Algorithm~\ref{algo:informal_algo} (below). The models specialize on different digits and styles.}\label{fig:comp_mixture}
\end{minipage}
\begin{minipage}{\textwidth}
\vspace{5mm}
\center\includegraphics[scale=0.13]{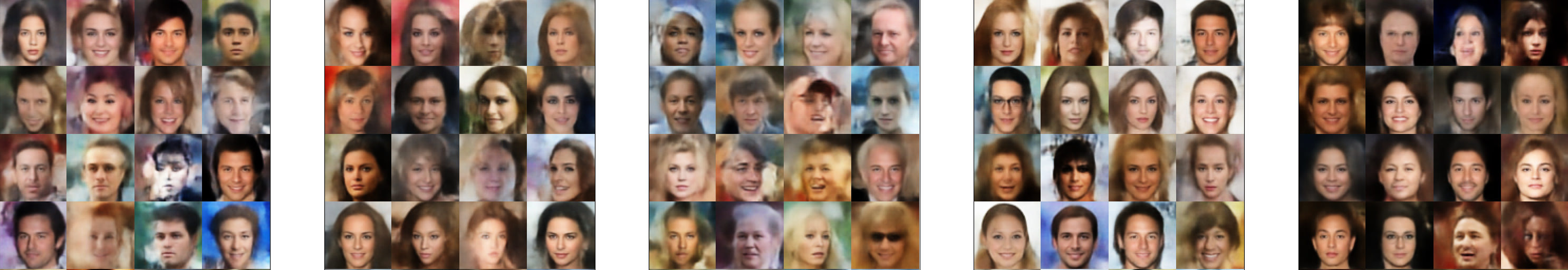}
\captionof{figure}{celebA: generated samples with VAEs separated per component after 26 iterations of Algorithm~\ref{algo:informal_algo}. The models specialize on different hair and background colors.}\label{fig:celeba_us}
\end{minipage}
\end{minipage}
\vspace{-2mm}
\end{figure*}

We generate synthetic data in two dimensions by first sampling 64,000 points from a mixture of Gaussian distributions and then we skew the second dimension $x_2$ with the non-linear transformation $x_2 = x_2 +  0.04 \cdot x_1^2 - 100 \cdot 0.04$. In this experiment we use VAEs. Since, they normally do not mode collapse we can show case the property of the training procedure (i.e. splitting is due to our technique and not mode collapse). We use a small and standard architecture for the VAE: a neural network with two hidden layers with 50 units each as both the decoder and the encoder. The discriminator has a similar architecture. We keep the size of the dataset and the architecture fixed and progressively increase the difficulty of the task by increasing the number of modes. In the first two experiments, each model perfectly covers a single mode even when a mode is significantly different than the others. With nine modes, some models are still trying to approximate multiple modes, yielding samples outside the support of the data distribution at the end of the training procedure. On the other hand, a single VAE completely fails at this task (see Appendix~\ref{app:rez_synt}). Finally, we compute the log-likelihood of the true data under the generated distribution using a Kernel Density Estimation with Gaussian kernel. We compare against a larger VAE with 150 (VAE-150) units per layer (instead of 50), trained uniformly over the training set, and a bag of VAEs with 50 units trained on a random subsample of the training set containing $N/K$ training points. We note that the random splitting of the training set did not help the VAE to specialize and actually made the log-likelihood worse after training for 100 epochs. Overall, our approach gives the best log-likelihood in this experiment as depicted in Table~\ref{tab:logl}.
\vspace{-0mm}
\subsection{MNIST and CelebA}
\begin{table}
\vspace{-0mm}
\caption{FID score on MNIST using VAEs (lower is better). }
\centering\begin{tabular}{ l  c  c  c  }
\toprule
  kVAEs & bag  & VAE-8  &VAE-64  \\
  \midrule
 9.99  &15.33  & 17.96 & 9.44	\\
 \bottomrule
\end{tabular}
\vspace{-0mm}
\end{table}
\looseness=-1For the experiment on MNIST, we do not know the number of modes. There is no reason to believe the optimal number of modes should be the number of digits. We arbitrarily use 15 models to capture the different strokes of the digits, following \citep{Tolstikhin:2017wo}.
We again use a small and simple architecture; the encoders and the decoders have 4 and 3 convolutional layers with 8-16-32-64 and 32-16-8 $4\times 4$ filters respectively. In Figure~\ref{fig:comp_mixture} we show the clustering of MNIST on which each VAE is trained and the corresponding generated digits.

\begin{figure*}[h!]
\vspace{-3mm}
  \centering
  \begin{subfigure}{0.6\textwidth}
    {\adjincludegraphics[scale=0.22, trim={7.3cm 62.6cm 6cm 10cm}, clip]{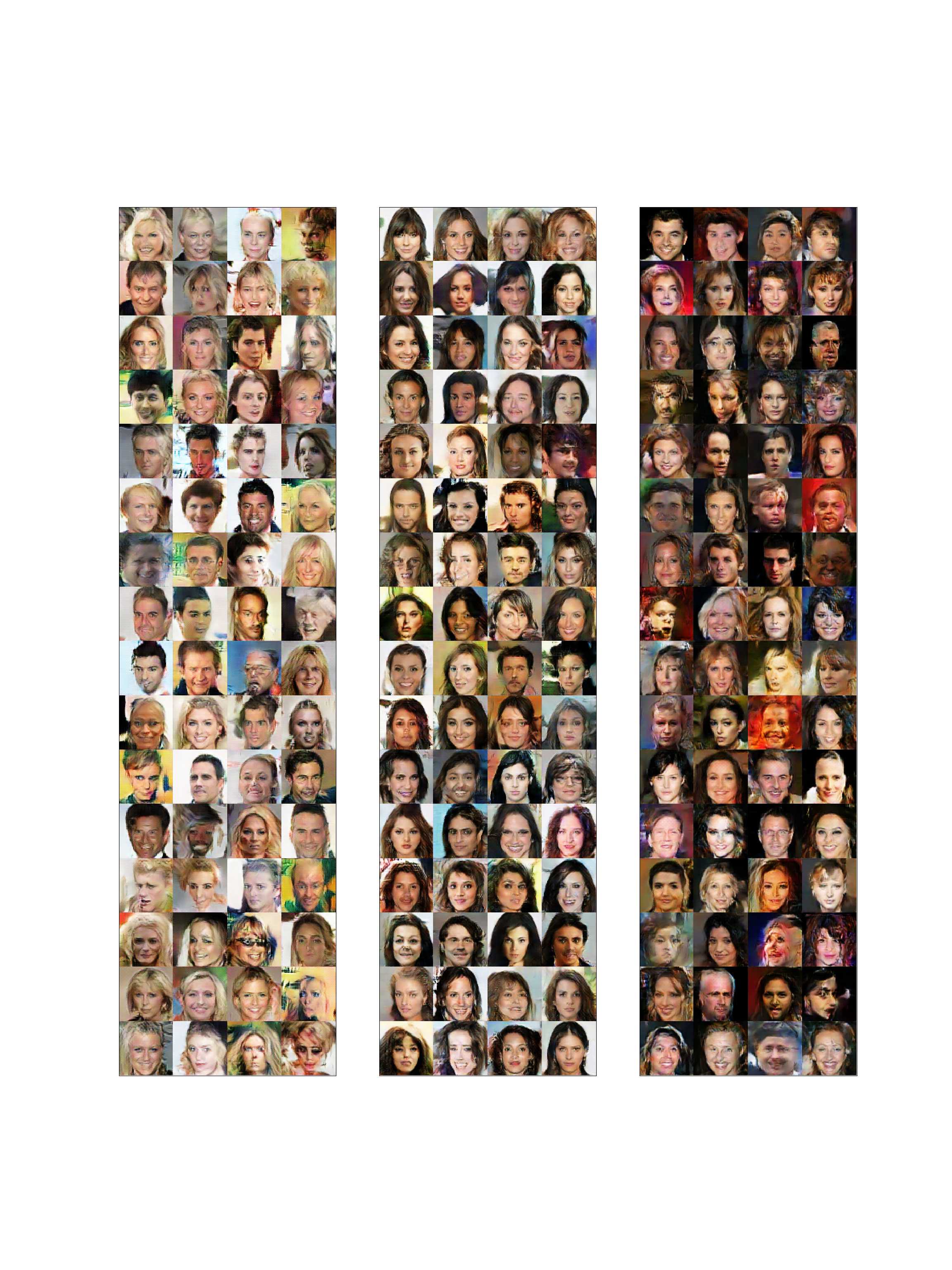}}
  \end{subfigure}%
  \begin{subfigure}{0.3\textwidth}
  \vspace{6mm}\hspace{0.3cm}
    {\adjincludegraphics[scale=0.22, trim={0.cm 65.6cm 0.cm 10cm}, clip]{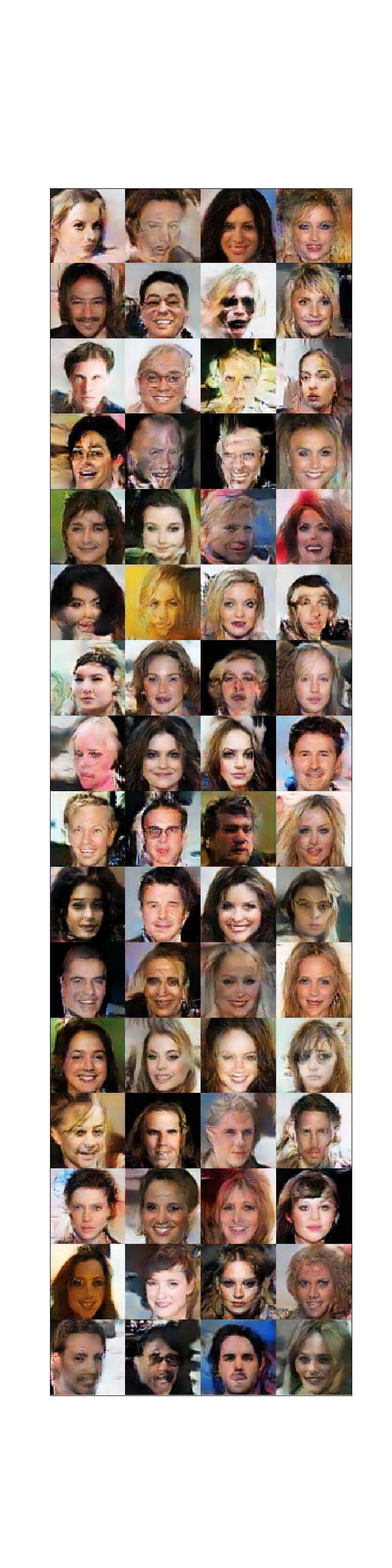}}
  \end{subfigure}
  \begin{subfigure}{0.6\textwidth}
    {\adjincludegraphics[scale=0.22, trim={7.3cm 62.6cm 6cm 10cm}, clip]{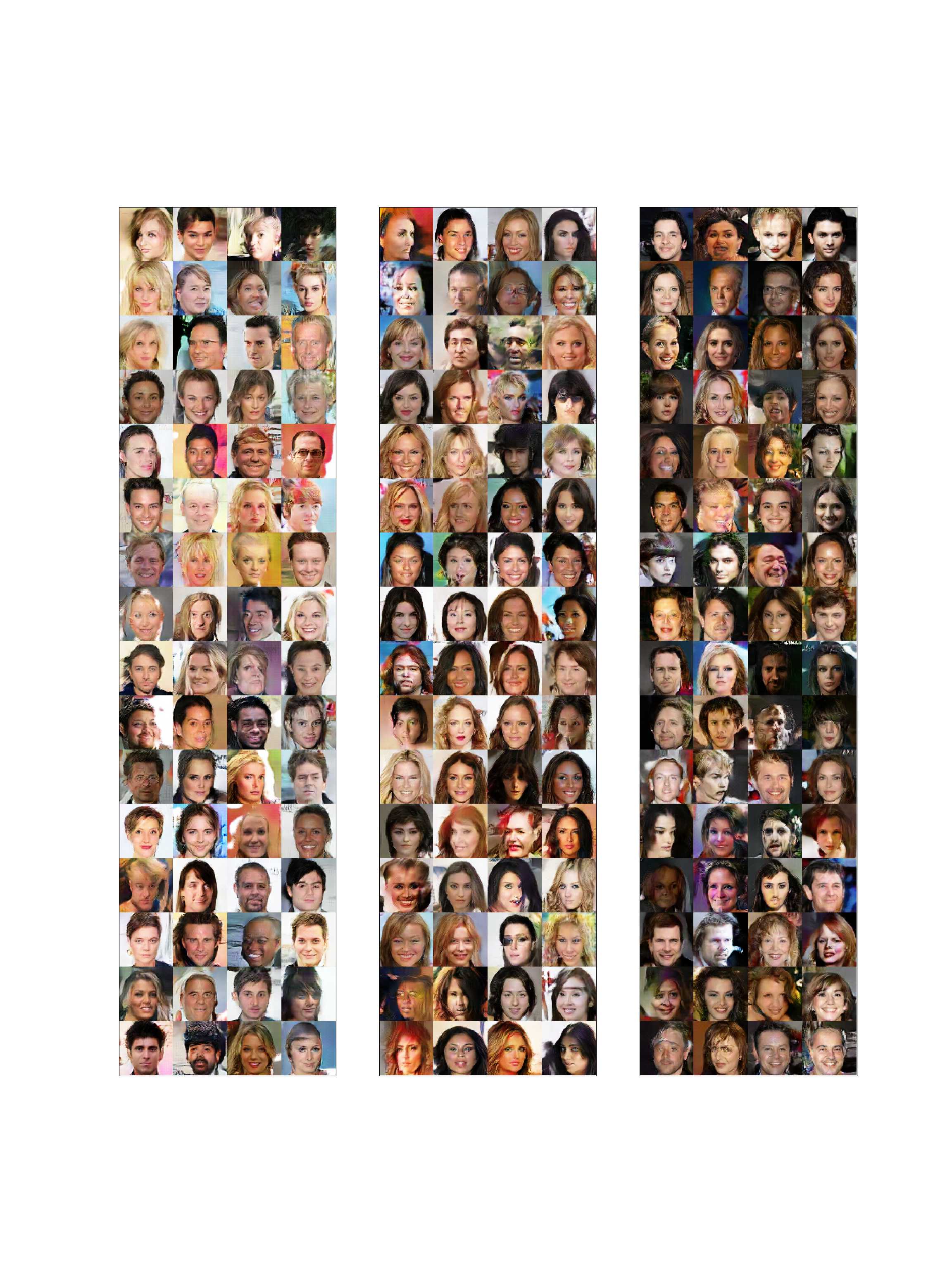}}
  \end{subfigure}%
  \begin{subfigure}{0.3\textwidth}
  \vspace{6mm}\hspace{0.3cm}
    {\adjincludegraphics[scale=0.22, trim={0.cm 65.6cm 0.cm 10cm}, clip]{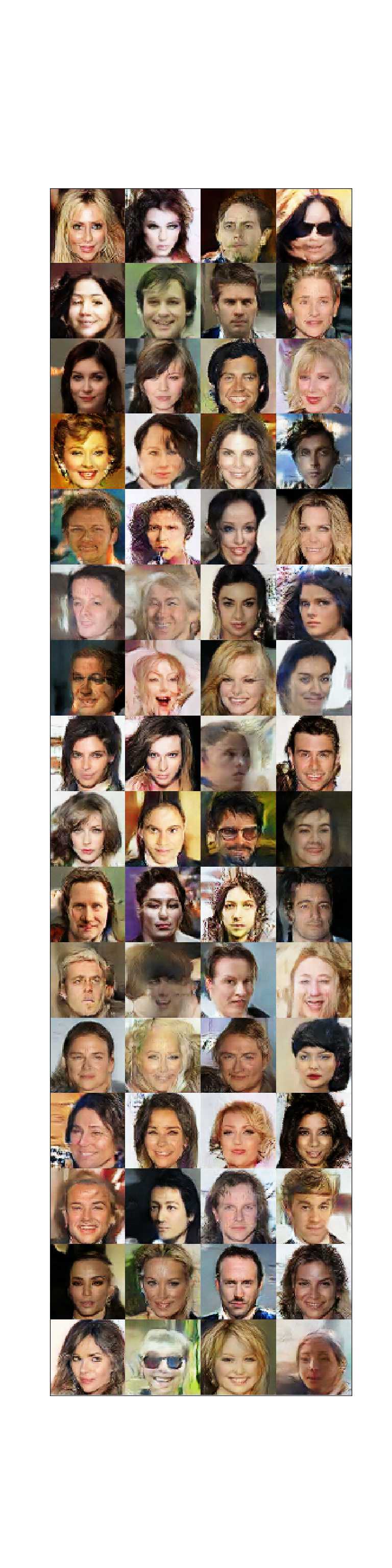}}
  \end{subfigure}
  \begin{subfigure}{0.6\textwidth}
    {\adjincludegraphics[scale=0.22, trim={7.3cm 62.6cm 6cm 10cm}, clip]{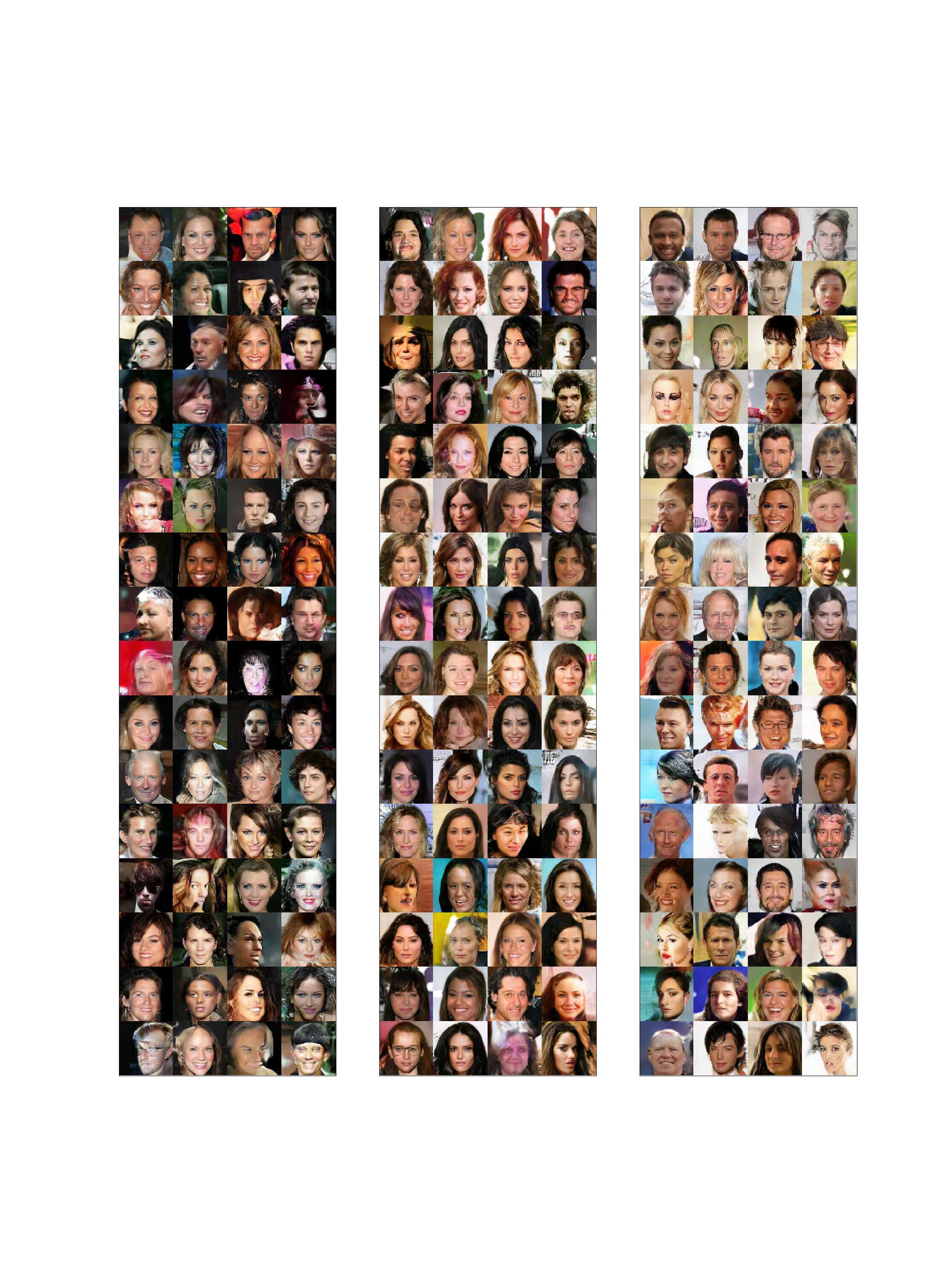}}
  \end{subfigure}%
  \begin{subfigure}{0.3\textwidth}
  \vspace{6mm}\hspace{0.3cm}
    {\adjincludegraphics[scale=0.22, trim={0.cm 65.6cm 0.cm 10cm}, clip]{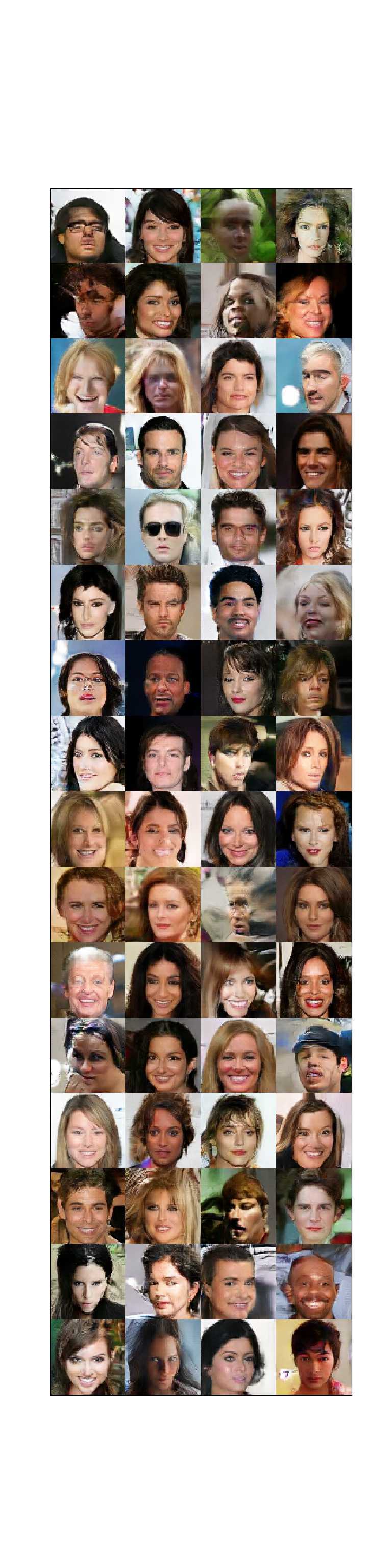}}
  \end{subfigure}
  \begin{subfigure}{0.6\textwidth}
    {\adjincludegraphics[scale=0.22, trim={7.3cm 62.6cm 6cm 10cm}, clip]{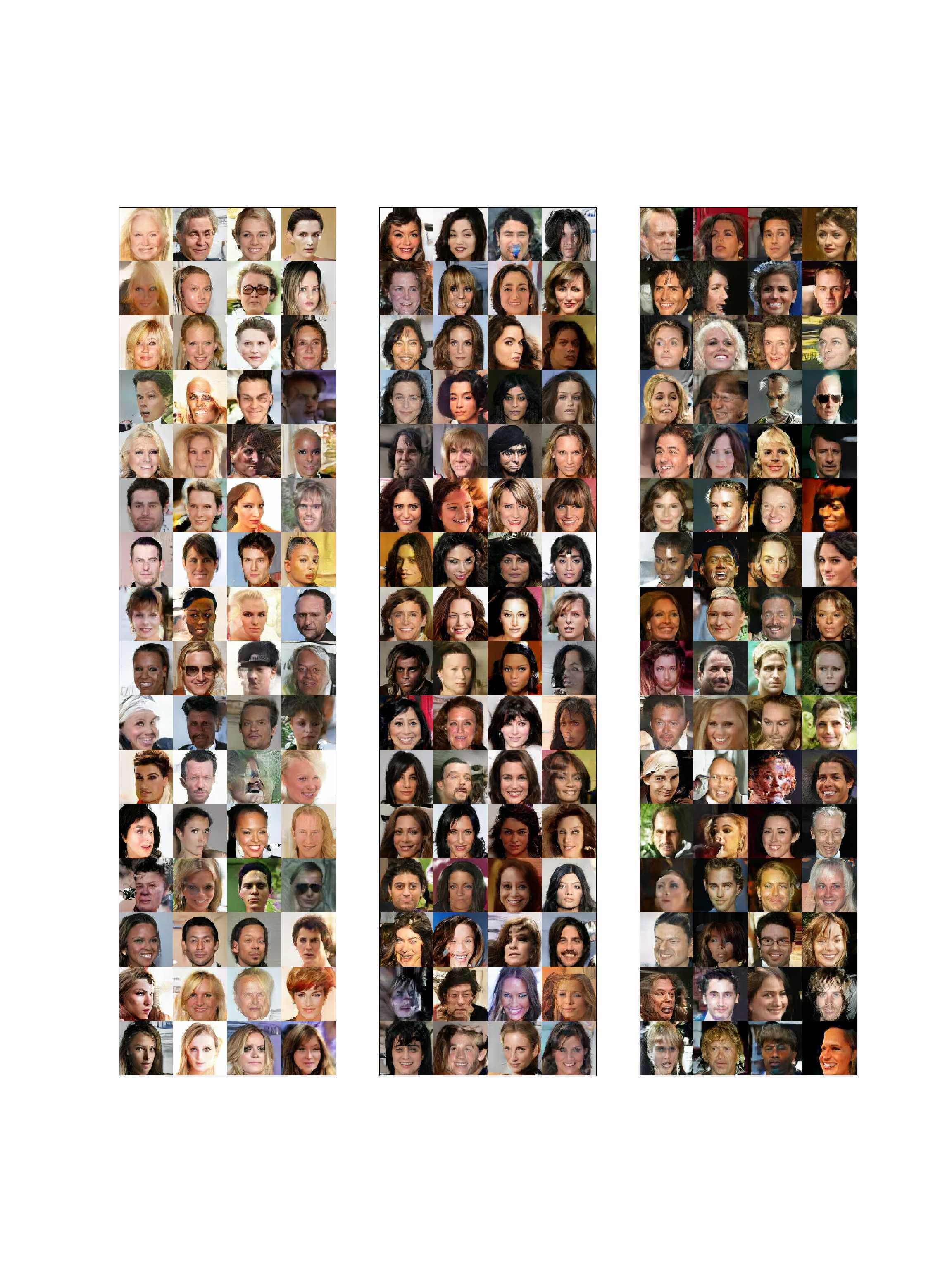}}
  \end{subfigure}%f
  \begin{subfigure}{0.3\textwidth}
  \vspace{6mm}\hspace{0.3cm}
    {\adjincludegraphics[scale=0.22, trim={0.cm 65.6cm 0.cm 10cm}, clip]{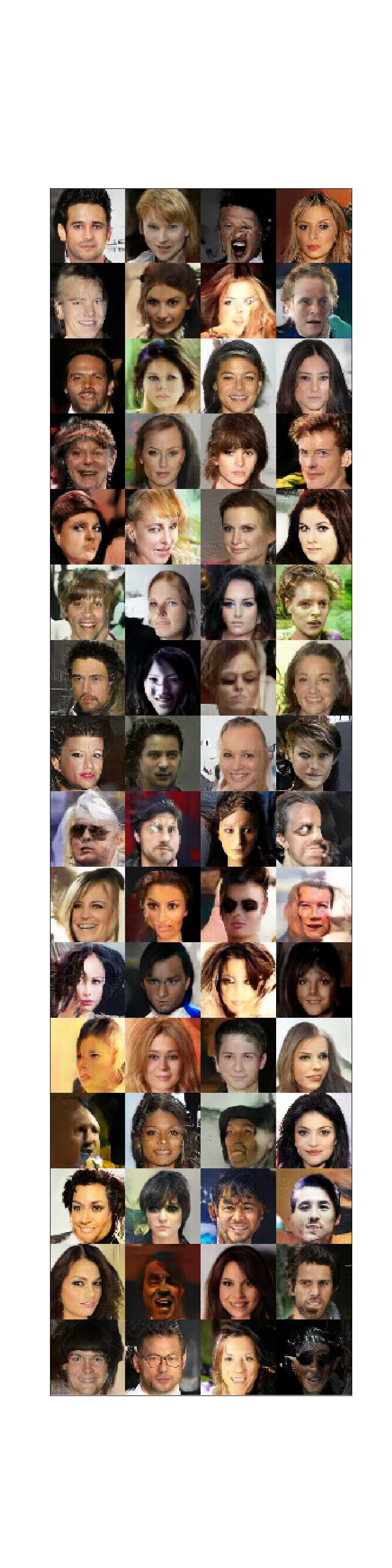}}
  \end{subfigure}
  \caption{Ablation study: 3 GANs (columns on the left) versus a single GAN (right column) trained on CelebA with increasing capacity (rows). The improvement upon a single model becomes marginal if the capacity of the generator is already high enough. \label{figure:GAN_ablation}}
\end{figure*}

Note that the different VAEs did specialize on distinct parts of the data distribution: similar digits tend to be grouped together (for example 3 and 8, 6 and 4, but also 5, 9 and 3 can be very similar depending on the style), as well as similar styles (tilted, thin, large, round, bold and combinations thereof). 
To evaluate our generated samples we used the FID score~\citep{heusel2017gans}. We remark that our FID score is competitive with the one obtained with a large VAE (VAE-64, 64-128-256-512 and 256-128-64 filters), as well as the one that can be obtained with GANs (according to Figure 5 of~\citep{lucic2017gans}, slightly below 10). It also is significantly better than both bagging and a single model with the same capacity. 

\begin{figure}
\vspace{-1mm}
\captionof{table}{FID score on celebA using VAEs.}\label{tab:FID_celeba}
\vspace{2mm}
\center\begin{tabular}{ l  c  c }
 \toprule
  kVAEs  & bag  & VAE-96\\
  \midrule
 \textbf{64.55}  & 71.90 & 67.06\\
 \bottomrule
\end{tabular}
\vspace{-3mm}
\end{figure}
\begin{figure}
\vspace{-2mm}
\captionof{table}{Ablation study on generator capacity: FID score on celebA using GANs.}\label{tab:ablationFID_celeba}
\center\begin{tabular}{ l  c  c  c c }
 \toprule
    & 64  & 128& 256&512\\
  \midrule
 3-GANs & \textbf{55.61}  & \textbf{31.64} & 23.75 & 19.38\\
 single GAN& 64.16 & 40.03 & 23.63 & 21.18\\
 \bottomrule
\end{tabular}
\vspace{-2mm}
\end{figure}
Finally, we test our algorithm on celebA using 5 components. We use the same architecture we used for MNIST but with 64-128-256-512 filters for the encoder, 256-128-64 for the decoder and 128 dimensional latent space. We note that the generated samples are visually clustered in Figure~\ref{fig:celeba_us}. We note that the background and the hair color plays a significant role in the clustering (see for example cluster 3 and 5). Again, our procedure improves the FID score as depicted in Table~\ref{tab:FID_celeba} which are competitive with the ones reported in Figure 5 of~\citep{lucic2017gans} and better than both bagging (same architecture as ours) and a single VAE with 96-192-384-768 and 384-192-96 filters for encoder and decoder respectively with latent space with 64 dimensions trained for 60 epochs.
\subsection{Generator Capacity}
\looseness=-1The previous experiments showed that the proposed algorithm reliably improve the FID score on VAEs. The competitive procedure relies on the fact that a single generator is not powerful enough to model the whole distribution. In this setting, splitting is therefore necessary in order to achieve a good quality in the generated samples. We now perform an ablation study on the capacity of the generator using GANs. 
We train a DC-GAN with spectral normalization~\citep{miyato2018spectral} and fixed hyper-parameters. We vary the maximum number of filters in the range $[64,128,256, 512]$ and train 3 GANs with our algorithm comparing with the FID of a single GAN of the same size. We train the models until convergence. Results are depicted in Table~\ref{tab:ablationFID_celeba}. From this experiment, we conclude that if the generative model is not very powerful, training multiple models substantially improves the FID. On the other hand, there is little to no gain when a single generative model alone  already generates good pictures.

\vspace{0mm}
\section{Conclusions}
\vspace{0mm}
In this paper, we introduce a clustering procedure using implicit generative models, which encourages them to generate more realistic samples. Our approach is inspired by the belief that the causal decomposition of a generative model into independent modules tends to yield simple components. Therefore, we train networks of limited capacity, competing with each other in the pursuit of generating more realistic samples. We enforce the competition between the models by relying on a set of discriminators that judge the quality of the samples produced by each model. We demonstrate how to decouple the loss of the combined model into parts that can be trained independently and show that our approach is a generalization of classical k-means clustering. 
We empirically validate that the model can successfully recover the true generative mechanisms. Even when recovery is not perfect, the competitive procedure allows to generate samples which are closer to the support of the data distribution. 
The approach we presented is modular and there are several possible extensions. Given enough computational resources, one could substantially increase the number of generative models. Furthermore, it can be extended by a better latent manifold structure, for example using Wasserstein autoencoders~\citep{tolstikhin2017wasserstein}. Benchmarking the multiple discriminators with a single K+1 discriminator is another interesting direction, as well as introducing parameter sharing between the generative models (conditional GANs~\citep{mirza2014conditional} where the label is given by the discriminators).
Finally, adaptively selecting the number of components does not have a trivial solution. Exploiting the vast clustering literature could shed some light on how to perform model selection on the fly. 
\newpage
\section*{Acknowledgement}
FL is partially supported by the Max-Planck ETH Center for Learning Systems and by an ETH core grant (to GR). This work was done while IT was at MPI.

\bibliography{bibliography}
\bibliographystyle{plainnat}
\newpage
\clearpage

\appendix
\onecolumn
\section{Proof of Lemma~\ref{lemma:parallel_training}}\label{lemma_app:parallel_training}
Trivially, all of the below hold from the definition of $c_j$:
\begin{itemize}
\item $\int_{\sfX^j} d\Pdj (x) = 1$ where $\sfX^j$ is the support of $d\Pd^j$
\item $\int_\sfX d\Pdj (x) = 1$ as $d\Pdj (x)$ is zero outside its support
\item $\cup_{j\in [K]} \sfX^j = \sfX$ and $\sfX^j \cap \sfX^{l}  = \emptyset$ $\forall$ $l\neq j$
\end{itemize}
By definition of the model we write the $f$-divergence as:
\begin{align*}
D_f(\Pm\|\Pd) &= D_f(\sum_{j=1}^k\alpha_jP_{g_j}\|\Pd)
\end{align*}
Now, we have that $\alpha_j = \int_\sfX d\Pd (x)c_j^{(t)}(x)$. Since $\sfX^j\cap \sfX^k = \emptyset$ for $j\neq k$, we can write:
\begin{align*}
D_f(\sum_{j=1}^k\alpha_jP_{g_j}\|\Pd) = D_f(\sum_{j=1}^k\alpha_jP_{g_j}\|\sum_{j=1}^k\alpha_j\Pdj^{(t)})
\end{align*}
 Joint convexity of $D_f$ concludes the proof~\cite{nowozin2016f}.

\section{Proof of Lemma~\ref{lemma:k-means}}
These arguments are trivial and can be found in any machine learning textbook. They are repeated here for completeness.
We now show that k-means clustering is a special case of our framework.
Assume that the data is generated by a mixture of Gaussians. We can lower bound the log-likelihood of the data using a variational bound:
\begin{align*}
\log(P(\cD_X)) \geq \sum_{i=1}^N \sum_{j=1}^K q_i(j) \log\left( \frac{P(x_i,j)}{q_i(j)}\right)
\end{align*}
 where $q$ is the variational approximation of the posterior and $j = 1,\ldots, K$ is the index of the components. One can then simply rewrite $P(x_i,j) = P(x_i|j)p(j)$. Then, for a Gaussian mixture model one parametrizes $P(x_i|j)$ with a Gaussian distribution. 
 
If the Gaussian is isotropic with vanishing covariance, the variational approximation of the posterior $q_i(j)$ degenerates to a hard assignment. Instead of approximating the generative model with a Gaussian distribution, we parametrize $P(x_i|j)$ with an implicit generative model from which it is easy to sample, i.e. the decoder of a VAE marginalized over the prior. Note that VAEs are trained to maximize the log-likelihood as in EM. Assume we have a Gaussian encoder which maps all the input to a single point (degenerate Gaussian with $\sigma=0$ and constant mean $\mu$ independent from the input $x$). Now, say we have the identity as decoder. Then, training the autoencoder amounts to minimizing:
 \begin{align*}
\min_{\mu_j} \bbE_{x\sim \Pdj} \left[ - \log P_{g_j}(x|\mu_j)\right] = \min_{\mu_j}\bbE_{x\sim \Pdj} \left[\frac{1}{2}\|x - \mu_j \|^2\right].
 \end{align*}
 Then, using EM, we compute the update for the (degenerate) variational distribution:
 \begin{align*}
 q_i(j = 1) = \lim_{\sigma\rightarrow 0} \frac{\alpha e^{-\|x_i-\mu_j\|/2\sigma}}{\sum_j\alpha_je^{-\|x_i-\mu_j\|/2\sigma}}
 \end{align*}
And recalling that $\log P_{g_j} = - \| x_i -\mu_j \|^2/2$ we notice that the degenerate posterior is obtained by maximizing the likelihood. In our approach, we estimate $P_{g_j}$ using a discriminator to account for the fact that we might not have a clear notion of distance. 
In an euclidean space, one can simply use a nearest neighbor classifier between the output of the VAEs (i.e., the centroids) and the training points. Note that this procedure is exactly k-means.
\section{Synthetic Data: Additional Results}\label{app:rez_synt}
We use a small and standard architecture for the VAE: a neural network with two hidden layers with 50 units each as both the decoder and the encoder. The discriminator has a similar architecture.
We use a 5 dimensional latent space and assume a Gaussian encoder. At each iteration, we train each VAE for 10 epochs on a split of the dataset (VAEs are pretrained uniformly on the dataset), and the classifier is trained for 2 epochs for the first two experiments. We use Adam~\cite{kingma2014adam} with step size 0.005, $\beta = 0.5$ batch size 32. 
The first two experiments contains 3 and 5 separate modes respectively and are depicted in Figure~\ref{pic:app_gmm3} and~\ref{pic:app_gmm5}. In the latter, we have one mode which is more complex than the others. In both cases, each model perfectly covers a single mode. This experiment illustrates that VAEs can learn how to generate samples in the support of the data, provided that are only asked to capture a sufficiently simple distribution.

In the next experiment, we consider 9 modes, each containing significantly fewer points than the previous ones. We first train each VAE for 1000 epochs because this task is significantly harder. We note that even after long training, the models perform poorly, as the data is too complex for such simple models. Then, we run our algorithm on the pre-trained models and each model is re-trained for 10 epochs on the split given by the classifiers. Remarkably, after only 10 iterations of Algorithm~\ref{algo:informal_algo}, the generators split to cover only limited parts of the data distribution, as one can see in Figure~\ref{pic:app_gmm9}. 
\begin{figure}
\vspace{-0.5mm}
\center\includegraphics[scale=0.25]{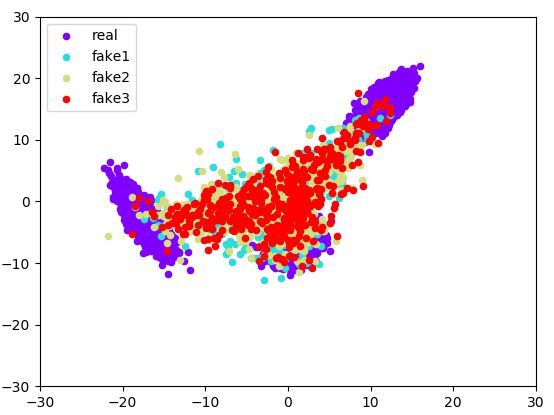}
\includegraphics[scale=0.25]{pics/gmm_3_100.jpg}
\caption{Synthetic data experiment, 3 modes and 3 VAEs after 10 epochs of uniform training and after 100 iterations of Algorithm~\ref{algo:informal_algo}.}\label{pic:app_gmm3}
\end{figure}

\begin{figure}
\center\includegraphics[scale=0.25]{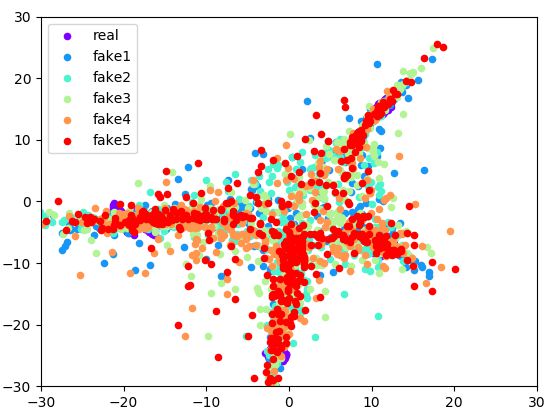}
\includegraphics[scale=0.25]{pics/gmm_5_100.jpg}
\caption{Synthetic data experiment, 5 modes and 5 VAEs after 100 epochs of uniform training and after 130 iterations of Algorithm~\ref{algo:informal_algo}.}\label{pic:app_gmm5}
\end{figure}
\begin{figure*}
\center\includegraphics[scale=0.21]{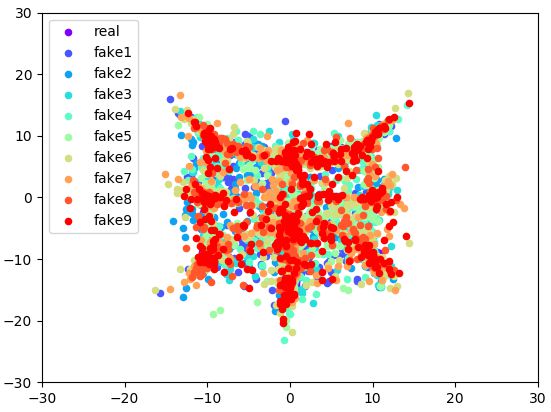}
\includegraphics[scale=0.21]{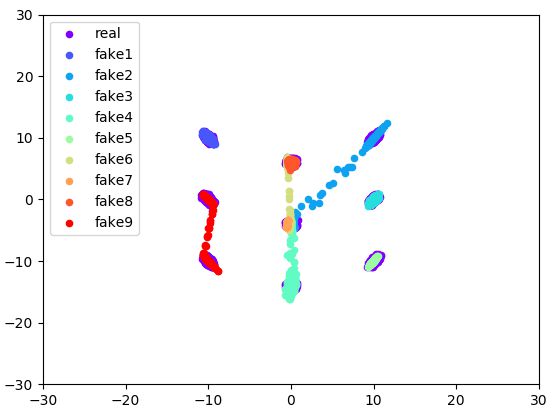}
\includegraphics[scale=0.21]{pics/gmm_9_100.jpg}
\caption{Synthetic data experiment, 9 modes and 9 VAEs after 1000 epochs of uniform training and after 10 and 100 iterations of Algorithm~\ref{algo:informal_algo}.}\label{pic:app_gmm9}
\end{figure*}

\section{MNIST: Additional Results}
We again use a small and simple architecture. The encoders and the decoders have 4 and 3 convolutional layers with 8-16-32-64 and 64-32-16 $4\times 4$ filters respectively. We use batch normalization with $\epsilon=10^{-5}$ and decay $0.9$. Each VAE has latent space dimension $8$, and we fix the learning rate of Adam for all networks to $0.005$. 
The discriminator has 3 convolutional layers and a linear layer with number of filters 64-128-256. As opposed to the synthetic data example, we do not reinitialize the classifier at each iteration of the meta algorithm. The reason is that we found the classifier output to be too sensitive to the initialization if it is not trained sufficiently long. On the other hand, training a full discriminator in every iteration was too expensive, and if trained too much, it would learn to distinguish fake example by just looking at specific blurriness patterns. 
In the synthetic experiments, the data produced by each VAE was indistinguishable from the real data if the support was correct, so training a classifier from scratch was feasible and gave best results.

In Figure~\ref{fig:cluster_app} we show the number of training samples assigned to each model divided by digit for the kVAE algorithm with 15 VAEs. 
In Figure~\ref{fig:kvae_app} we show samples from our model trained for 10 iterations of Algorithm~\ref{algo:informal_algo}.
In Figure~\ref{fig:bagging_app} we show samples from the bagging mixture of 15 VAEs trained for 100 epochs. We use the same architecture as the one used for the kVAEs Algorithm. In Figure~\ref{fig:mixture_single} and~\ref{fig:mixture_single512} we show samples from a single small and large VAE. We notice that our model produces a large variety of different styles which are visually clustered. 

\begin{figure*}
\center\includegraphics[scale=0.2]{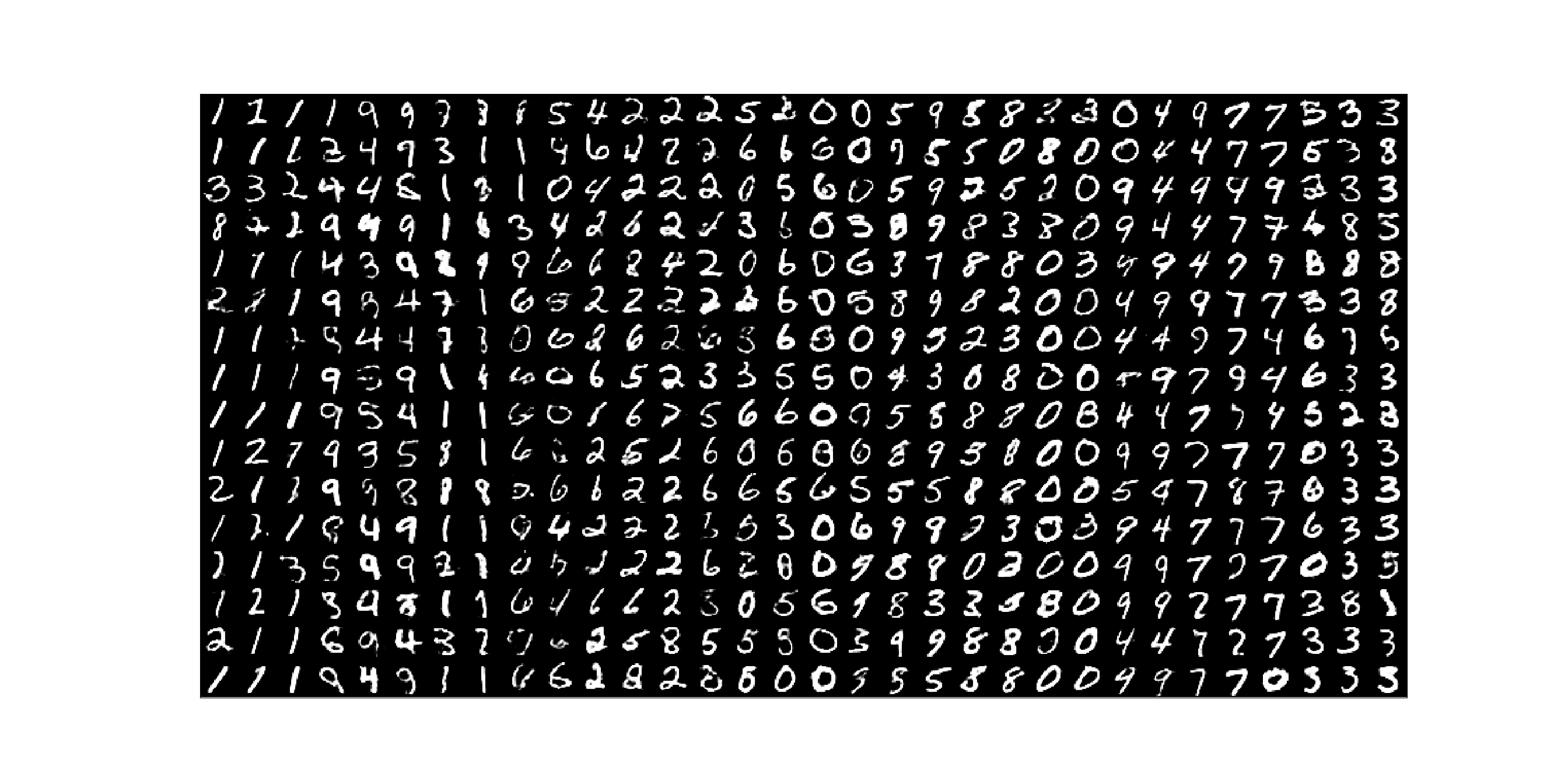}
\caption{Samples from the mixtures of kVAE with 15 components trained for 10 iterations of Algorithm~\ref{algo:informal_algo}. We notice a large variety of different strokes and styles.}\label{fig:kvae_app}
\end{figure*}

\begin{figure*}
\center\includegraphics[scale=0.1]{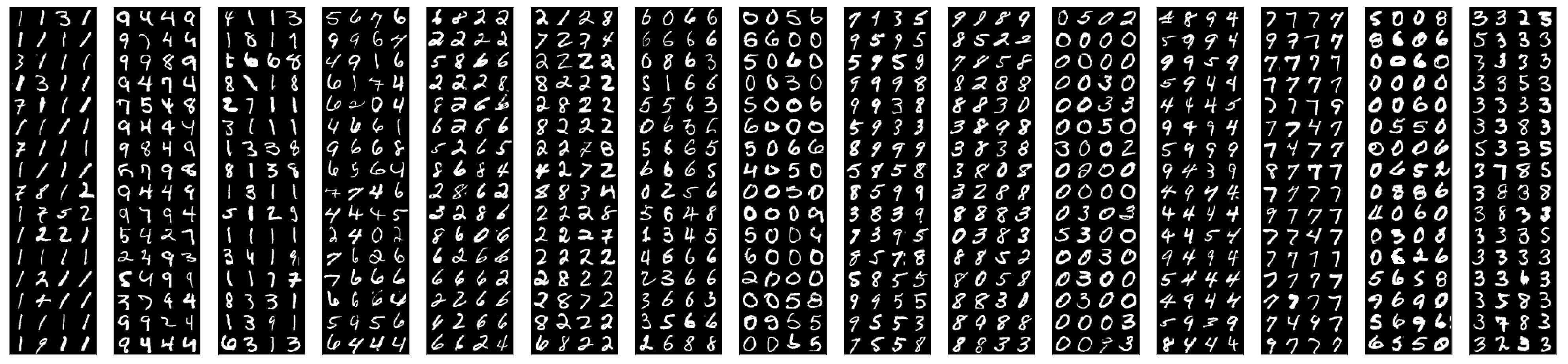}
\caption{Clustering of MNIST using the discriminators.}\label{fig:cluster_app}
\end{figure*}

\begin{figure*}
\center\includegraphics[scale=0.2]{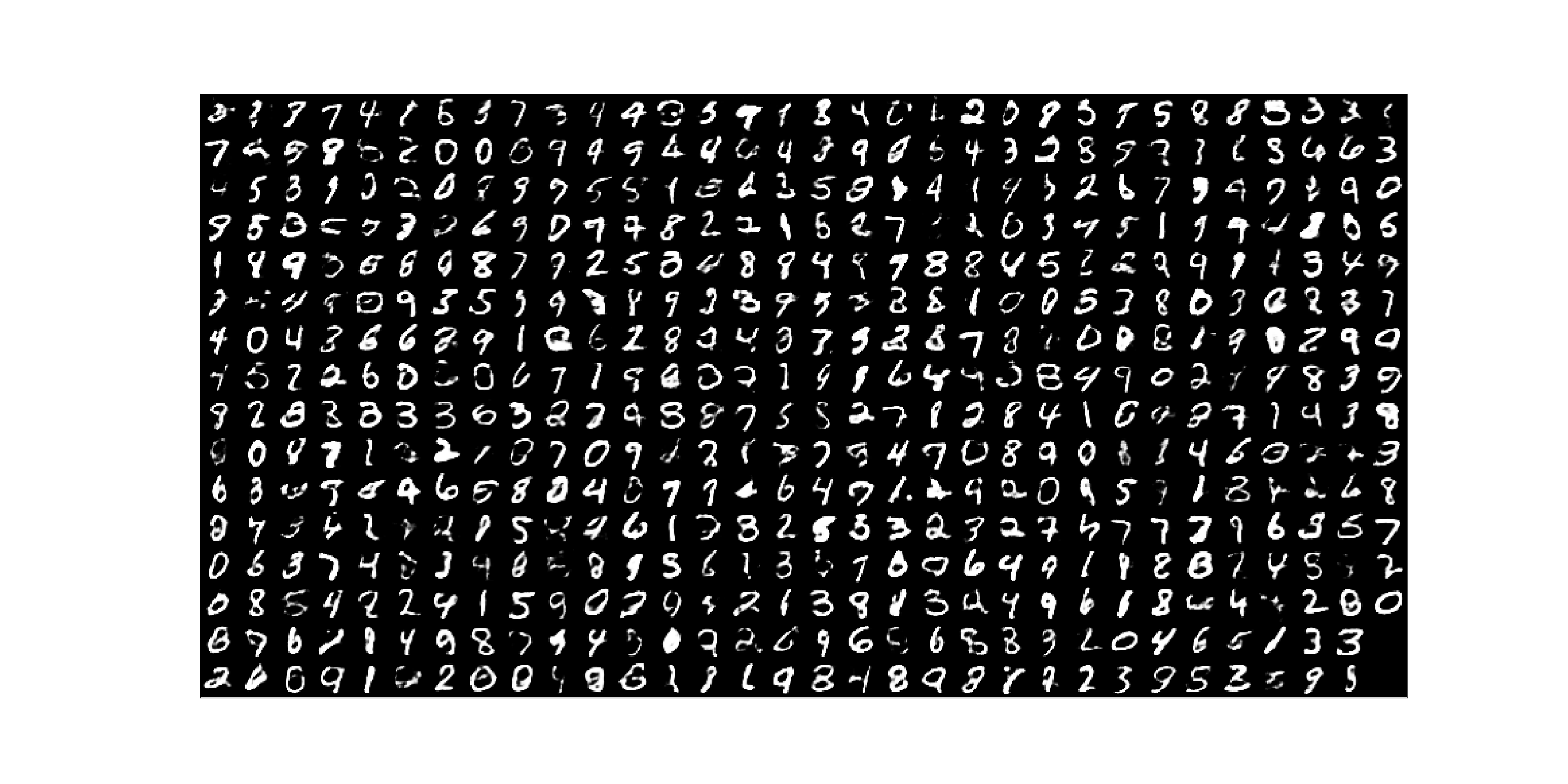}
\caption{Samples from the bagging of 15 VAEs with the same architecture as ours trained on random splits of the training data}\label{fig:bagging_app}
\end{figure*}

\begin{figure*}
\center\includegraphics[scale=0.2]{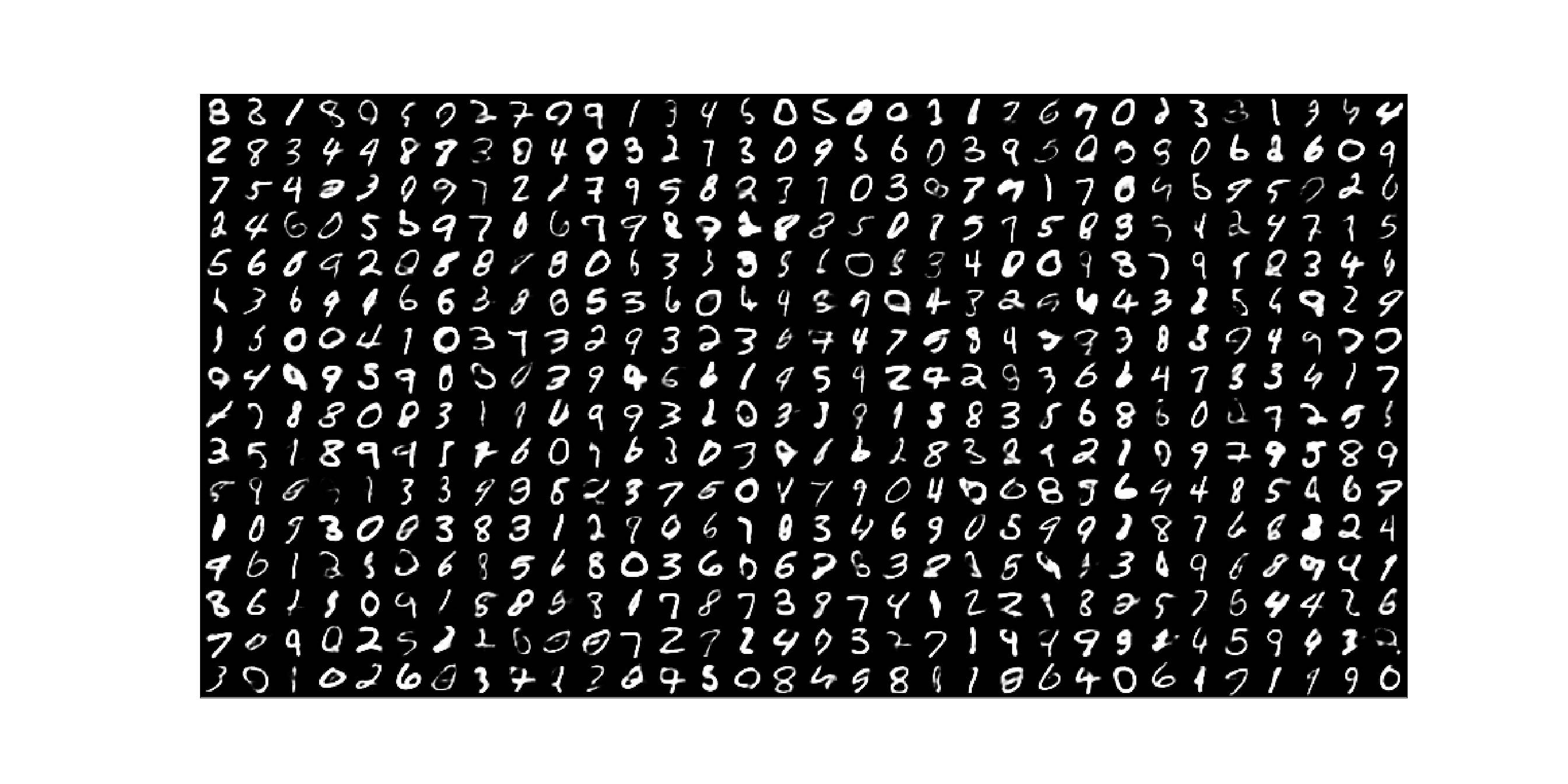}
\caption{Samples from a single VAE with the same architecture as ours}\label{fig:mixture_single}
\end{figure*}

\begin{figure*}
\center\includegraphics[scale=0.2]{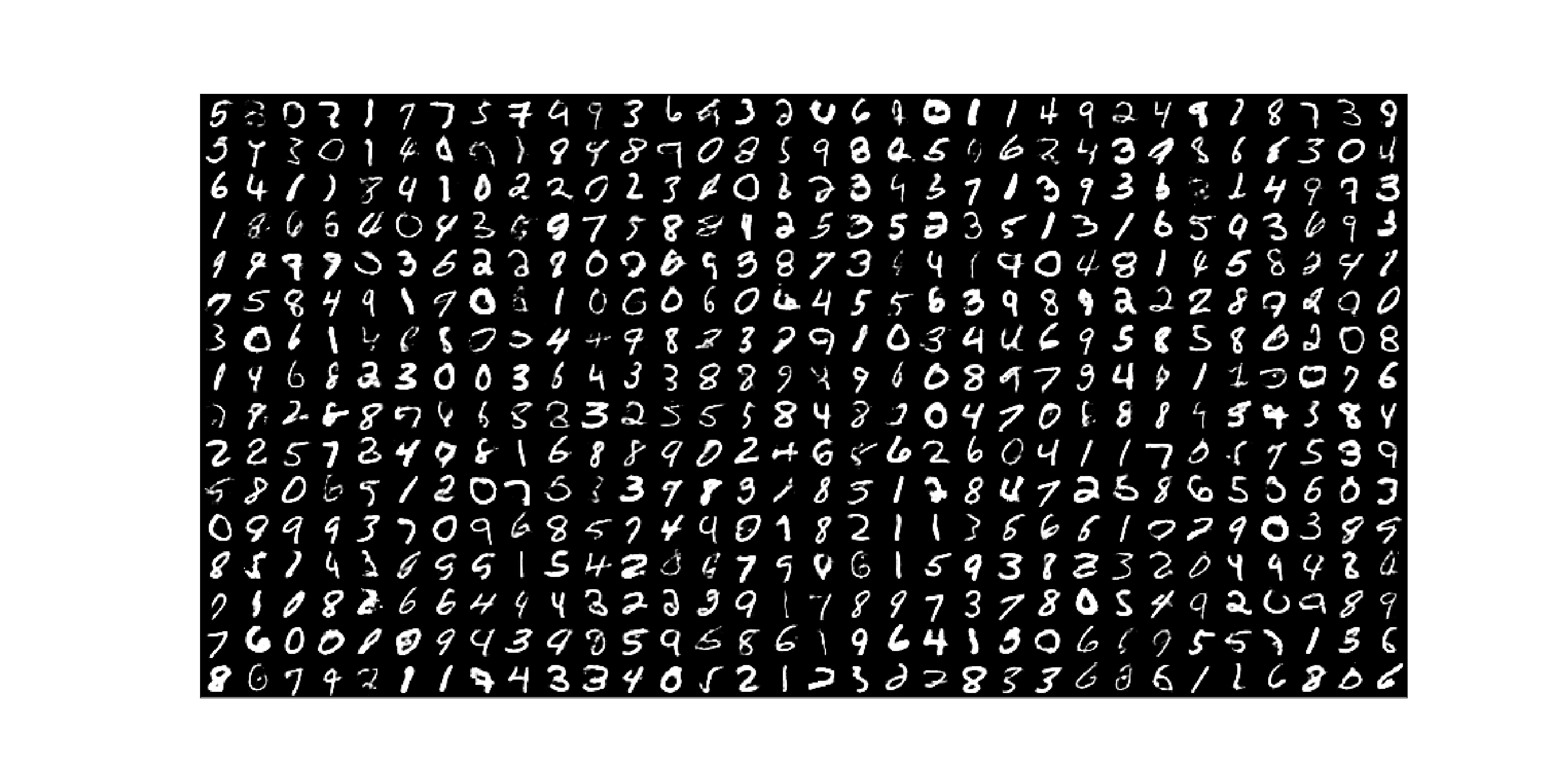}
\caption{Samples from a single VAE with 64-128-256-512 and 256-128-64 filters per layer }\label{fig:mixture_single512}
\end{figure*}
\newpage
\section{CelebA: Additional Results}
The encoders and the decoders have 4 and 3 convolutional layers with 64-128-256-512 and 256-128-64 $5\times 5$ filters respectively. We use batch normalization with $\epsilon=10^{-5}$ and decay $0.9$. Each VAE has latent space dimension 128, and we fix the learning rate of Adam to $0.0002$. The discriminator has 3 layers with 128 $3\times 3$ filters and is trained with SGD with stepsize $10^{-4}$. We perform an assignment after every 10 epochs of training. 
In Figure~\ref{fig:celebaD_bag_259_big} we show the samples from 5 models trained with bagging using the same architecture for each VAE and in Figure~\ref{fig:celebaD_us_259_big} the samples from our model. In figure~\ref{fig:celeba_single} we show samples from a VAE with 96-192-384-768 and 384-192-96 filters for encoder and decoder respectively with latent space with 64 dimensions trained for 60 epochs. We notice that our model produces more visually appealing and diverse samples than both bagging and a single larger model. For the ablation study using GANs, we use the same hyperparameters as we used for VAEs, except the learning rate of the GANs which is 0.0001.
\begin{figure*}
\center\includegraphics[scale=0.1]{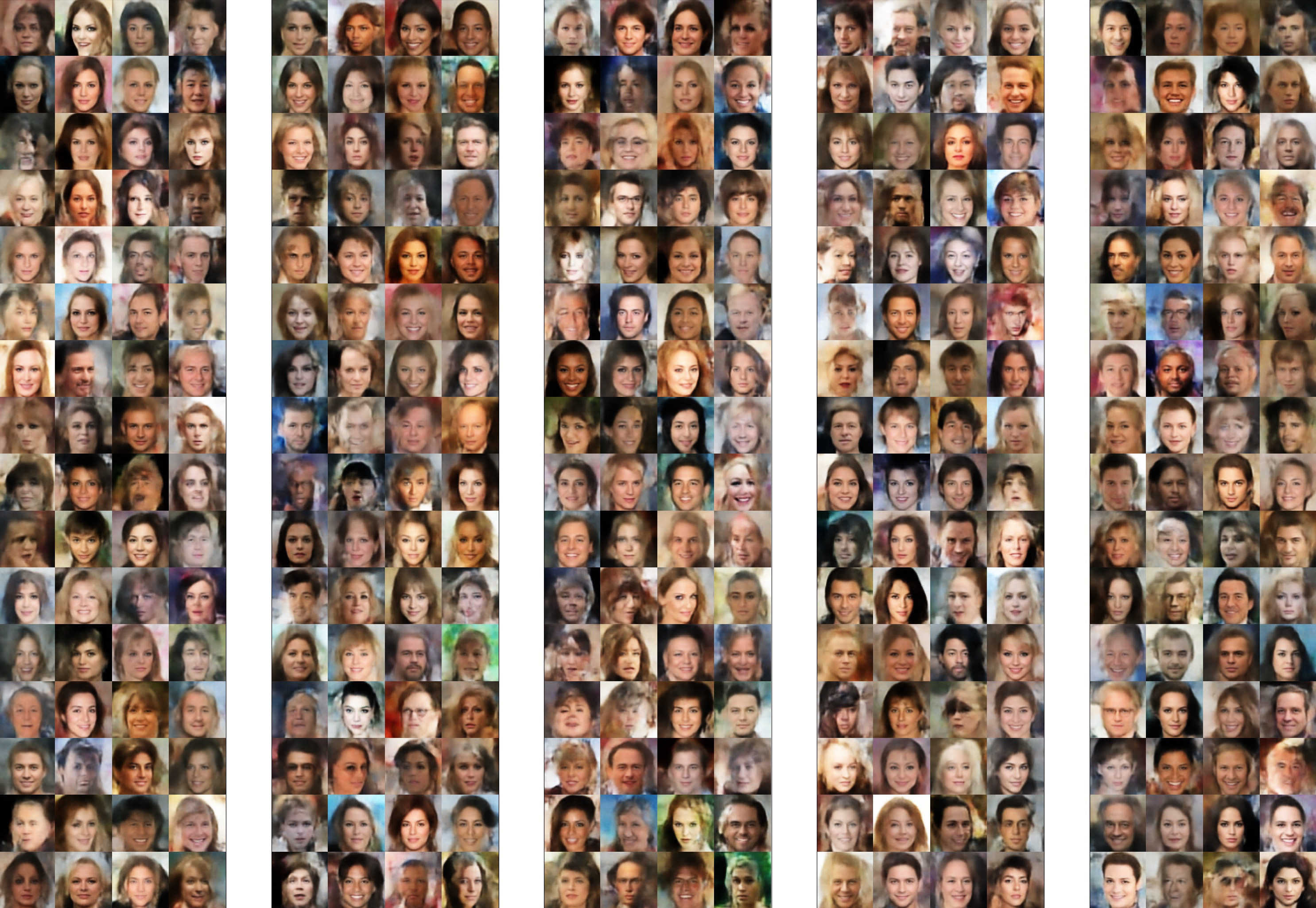}
\caption{samples from the bagging of 5 VAEs}\label{fig:celebaD_bag_259_big}
\end{figure*}

\begin{figure*}
\begin{minipage}{\textwidth}
\begin{minipage}{0.7\textwidth}
\center\includegraphics[scale=0.1]{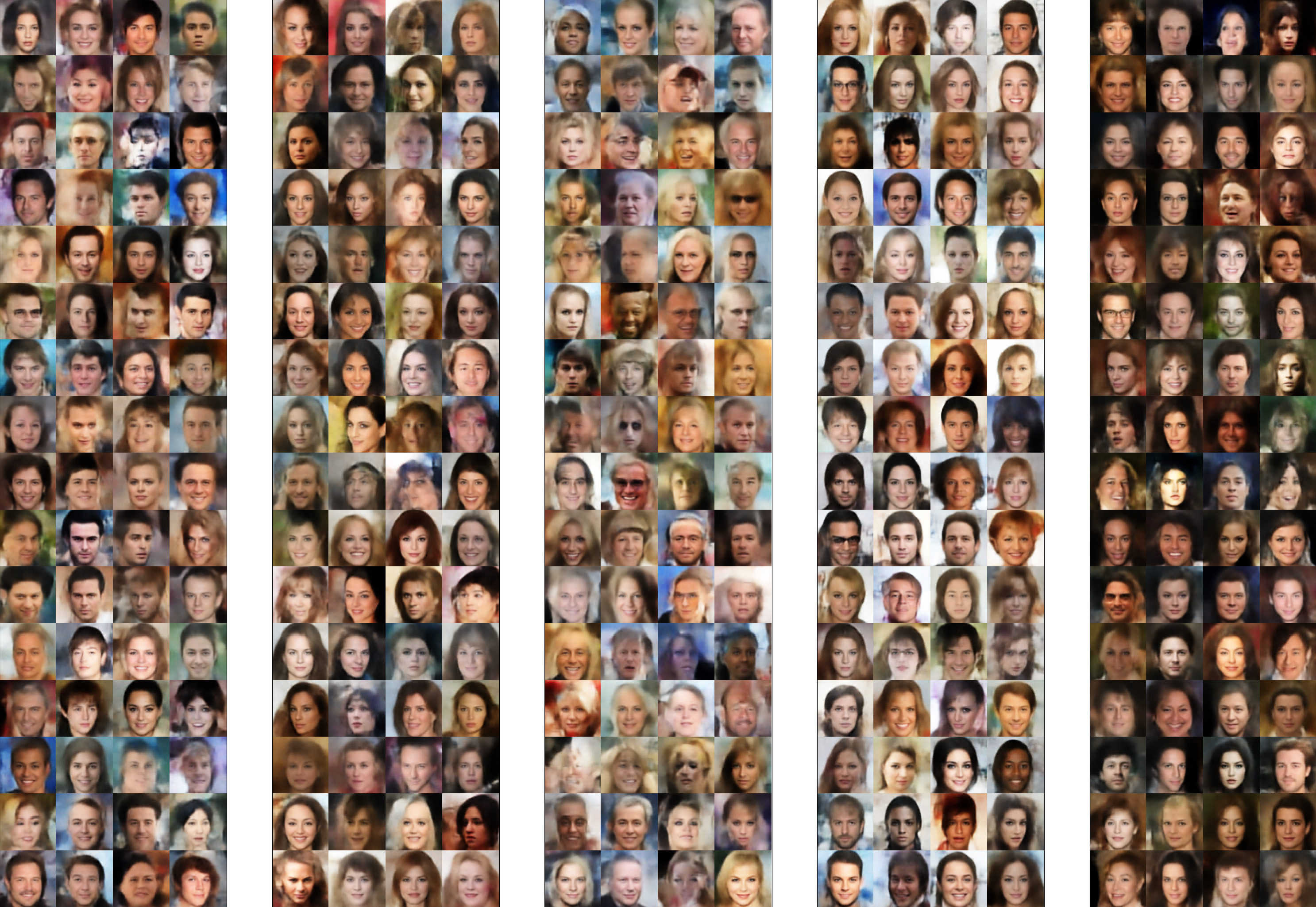}
\captionof{figure}{samples from 5 VAEs trained with our competitive procedure}\label{fig:celebaD_us_259_big}
\end{minipage}
\hspace{1cm}
\begin{minipage}{0.25\textwidth}
\center\includegraphics[scale=0.087]{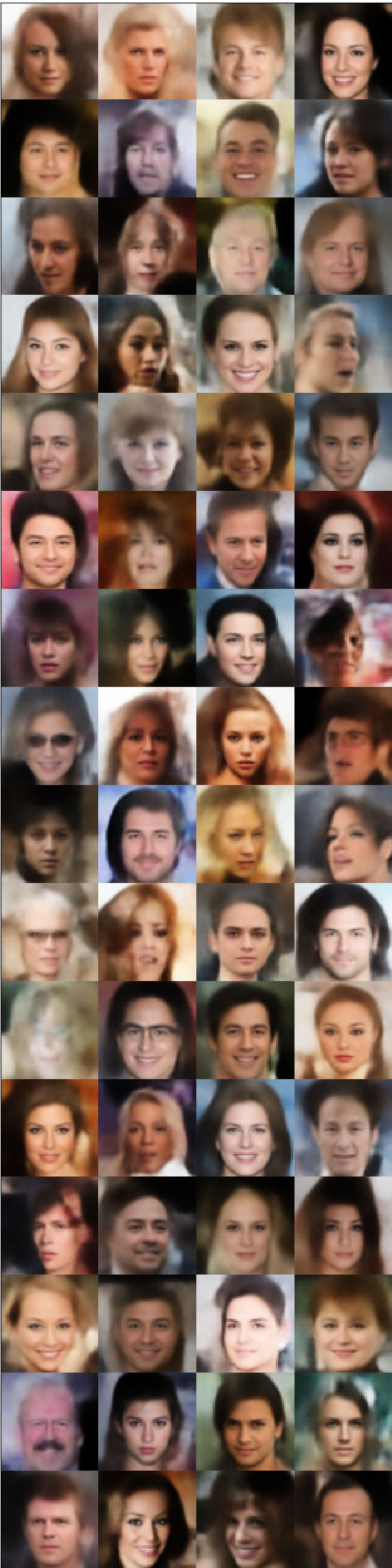}
\captionof{figure}{samples from a single larger VAE}\label{fig:celeba_single}
\end{minipage}
\end{minipage}
\end{figure*}

\end{document}